\documentclass[10pt,twocolumn,letterpaper]{article}

\usepackage{iccv}
\usepackage{times}
\usepackage{epsfig}
\usepackage{graphicx}
\usepackage{amsmath}
\usepackage{amssymb}

\usepackage{bibentry}
\usepackage{enumitem}
\usepackage{amsmath}
\usepackage{amsfonts}
\usepackage{amssymb}
\usepackage{makecell}
\usepackage{multirow}
\usepackage{graphicx}
\usepackage[table,xcdraw]{xcolor}
\usepackage{tikz}
\usepackage{color}
\usepackage{booktabs}

\newcommand{\Mmat}[0]{{{\bf M}}}
\newcommand{\Nmat}[0]{{{\bf N}}}

\newcommand{\Xmat}{{\bf X}}
\newcommand{\Ymat}[0]{{{\bf Y}}}

\newcommand{\fv}{\boldsymbol{f}}

\newcommand{\nv}{\boldsymbol{n}}

\newcommand{\pv}[0]{{\boldsymbol{p}}}

\newcommand{\vv}{\boldsymbol{v}}

\newcommand{\xv}{\boldsymbol{x}}
\newcommand{\yv}{\boldsymbol{y}}

\newcommand{\Phimat}{\boldsymbol{\Phi}}

\newcommand{\epsilonv}{\boldsymbol{\epsilon}}

\newcommand{\piv}{\boldsymbol{\pi}}

\newcommand{\tsp}{^{\top}}

\usepackage[pagebackref=true,breaklinks=true,letterpaper=true,colorlinks,bookmarks=false]{hyperref}

\iccvfinalcopy 

\def\iccvPaperID{5951} 

\ificcvfinal\fi

\begin{document}

\title{Unfolding Framework with Prior of Convolution-Transformer Mixture and Uncertainty Estimation for Video Snapshot Compressive Imaging}

\author{Siming Zheng\\
Computer Network Information Center,Chinese Academy of Science,\\
Beijing, 100190, China\\
University of Chinese Academy of Sciences,\\
Beijing 100049, China\\
{\tt\small zhengsiming@cnic.cn}
\and
Xin Yuan\\
School of Enginering, Westlake Universiy\\
Hangzhou, 310024, China\\
{\tt\small xylab@westlake.edu.cn}
}

\maketitle
\ificcvfinal\thispagestyle{empty}\fi

\begin{abstract}
We consider the problem of video snapshot compressive imaging (SCI), where sequential high-speed frames are modulated by different masks and captured by a single measurement. The underlying principle of reconstructing multi-frame images from only one single measurement is to solve an ill-posed problem. By combining optimization algorithms and neural networks, deep unfolding networks (DUNs) score tremendous achievements in solving inverse problems. In this paper, our proposed model is under the DUN framework and we propose a 3D Convolution-Transformer Mixture (CTM) module with a 3D efficient and scalable attention model plugged in, which helps fully learn the correlation between temporal and spatial dimensions by virtue of 
Transformer. To our best knowledge, this is the first time that Transformer is employed to video SCI reconstruction. Besides, to further investigate the high-frequency information during the reconstruction process which are neglected in previous studies, we introduce {\bf {\em variance estimation}} characterizing the {\bf  {\em uncertainty}} on a pixel-by-pixel basis. Extensive experimental results demonstrate that our proposed method achieves state-of-the-art (SOTA) (with a \textbf{1.2dB} gain in PSNR over previous SOTA algorithm) results. We will release the code.
\end{abstract}

\section{Introduction}
Nowadays, due to the ability of capturing high-dimensional data in an efficient way, Snapshot Compressive Imaging (SCI)~\cite{Llull13_OE_CACTI,Yuan2021_SPM} has attracted much attention. SCI system just employs a low-speed 2D camera to capture 3D sequential video frames, hyperspectral data, etc.~\cite{zheng2021super,Chen:22}, where a digital micro-mirror device~\cite{hitomi2011video,Reddy11_CVPR_P2C2} or a shifting mask~\cite{Yuan14CVPR} is utilized to modulate consequent frames. With the knowledge of modulation, the captured single 2D measurement can be reconstructed to original sequential frames by algorithms~\cite{iliadis2020deepbinarymask,Wu_2021_ICCV,li2020end,9428320,zheng2021deep, zhang2022herosnet,yoshida2018joint}. In this paper, we focus on the reconstruction problem of video SCI systems. 

\begin{figure}[t]
\centering
\includegraphics[width=.95\columnwidth]{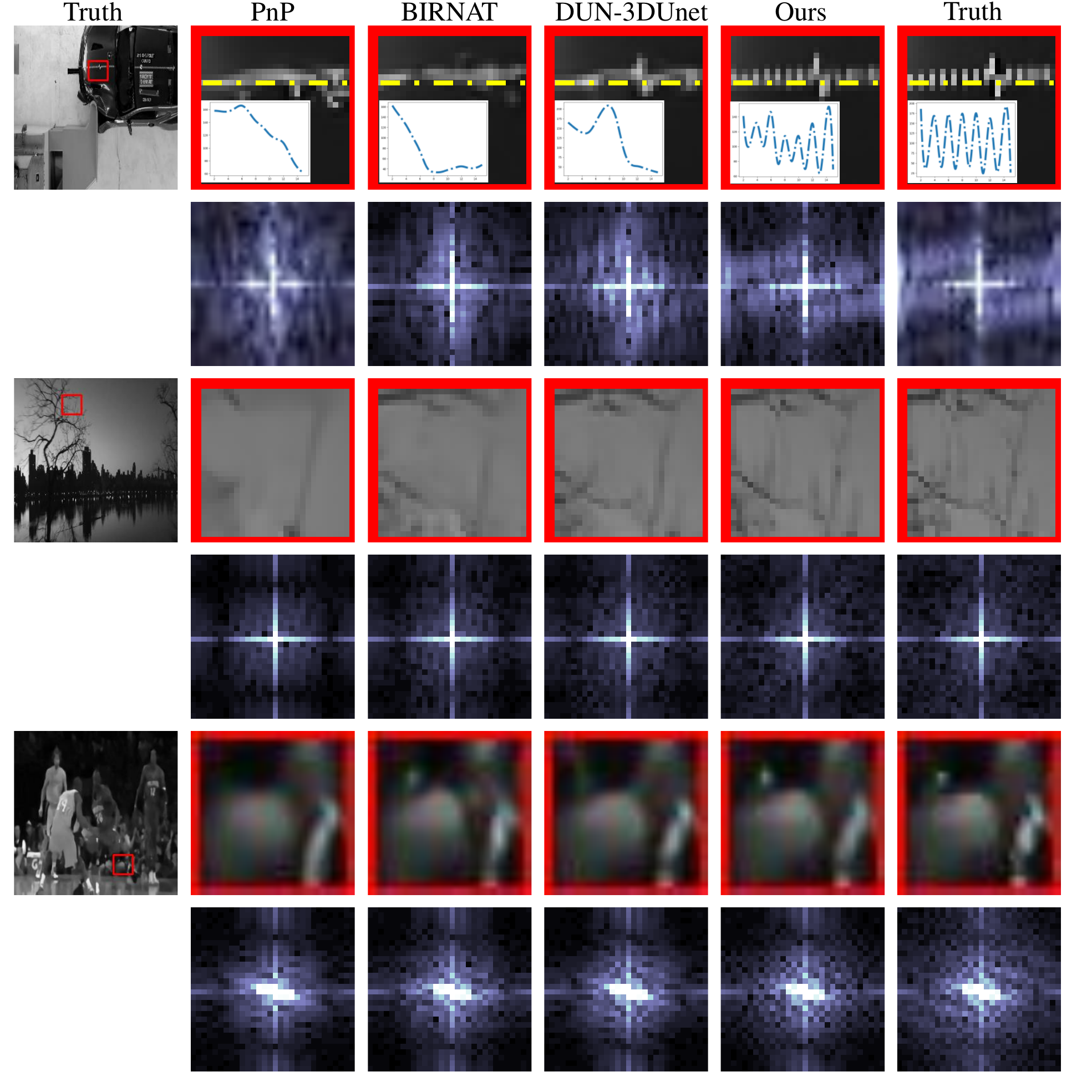}
\caption{{Illustration of comparison between the reconstruction results in the high-frequency details of previous SOTA algorithms and Ours. We present the details both in image domain (first line) and frequency domain (second line). To be more visually clearly, we also present the intensity profiles extracted from the cross-section yellow lines in the first example. Our proposed algorithm can reconstruct better high-frequency details.
}}
\vspace{-5mm}
\label{fig:teasing}
\end{figure}

To be concrete, the reconstruction process can be regard as solving an ill-posed inverse problem where the number of pixels to be reconstructed is much higher than the number of known parameters. 
%
With the development of deep learning, deep neural networks has been employed to conduct the reconstruction in recent years, where convolutional neural networks (CNN) are dominating. Compared to optimization-based algorithms, learning-based algorithms can directly map the measurement and the target images which makes it easier and faster to bring reconstruction results up to the mark. To improve learning-based algorithms' defect of lacking interpretability, recently proposed deep unfolding networks (DUNs) ~\cite{Wu_2021_ICCV,li2020end,9428320} combine the merits of both optimization-based and learning-based algorithms, and achieve the best results so far. 

\begin{figure}[t]
\centering
\includegraphics[width=1\columnwidth]{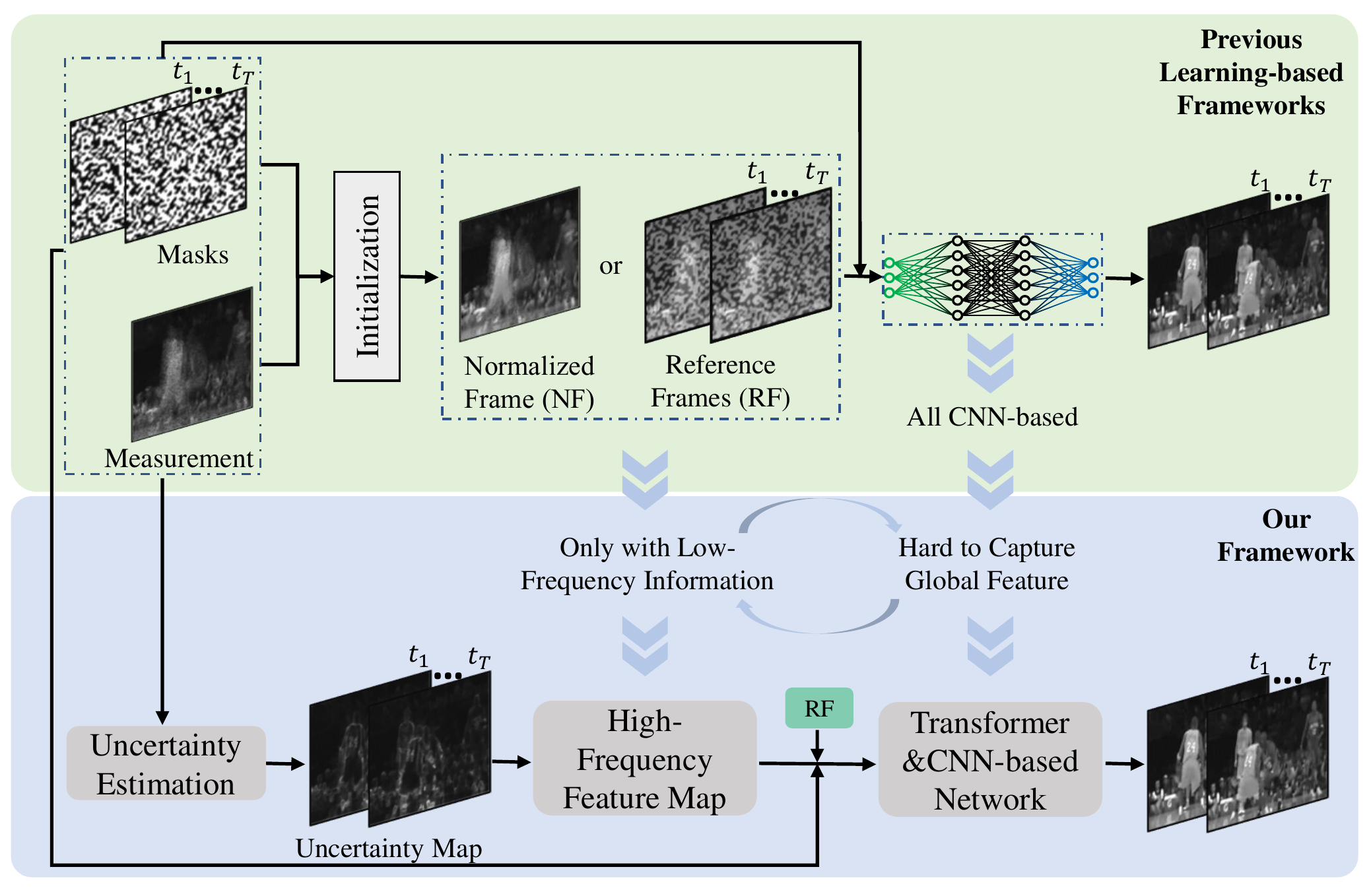}
\caption{{Illustration of the design motivation.}}
\vspace{-5mm}
\label{fig:motivation}
\end{figure}

\begin{figure*}[!htbp]
\centering
\includegraphics [width=2.05\columnwidth]{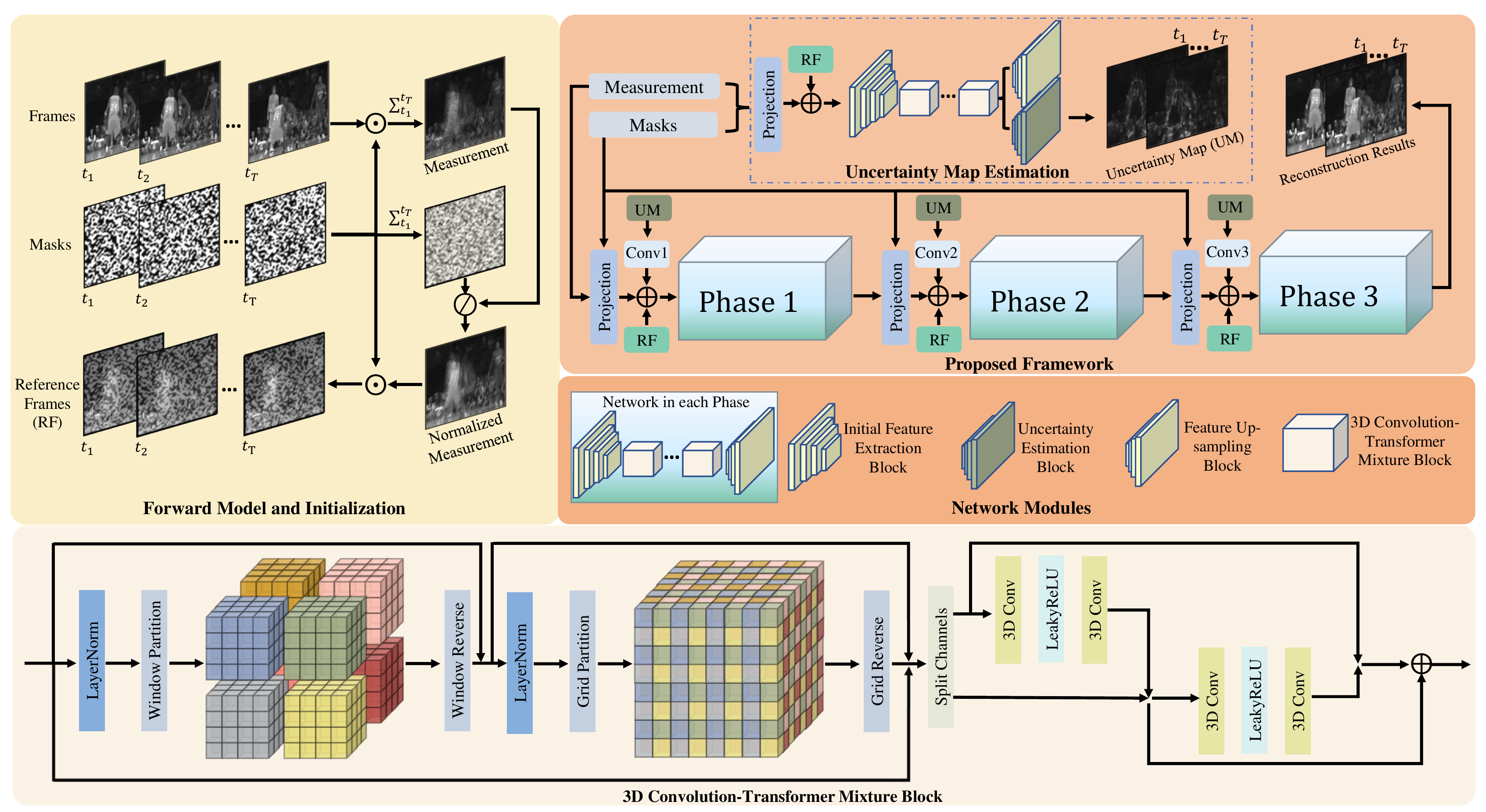}
\caption{ {Illustration of video SCI and our proposed model. Top-Left: Sequential video frames are modulated by dynamic masks and then compressed to the measurement. Normalized Measurement is achieved by element-divide the sum of dynamic masks. Reference Frames are acquired by element-wise multiplication. Top-Right: Architecture of our proposed uncertainty guided DUN for video SCI.
Bottom: Details of CTM blocks, composed of 3D scalabe blocked local and dilated global attention combining 3D-CNN. $\bigoplus$ here denotes concatenation. More details are in Supplementary Material (SM).}}
\vspace{-5mm}
\label{fig:Network}
\end{figure*}

\noindent{\bf {Motivation}:}
As shown in Fig.~\ref{fig:teasing}, previous SOTA learning-based algorithms ~\cite{Wu_2021_ICCV,cheng2020birnat,yuan2020plug} are not ideal for high-frequency detail reconstruction. In their initialization (details in Fig.~\ref{fig:Network}), as shown in the top-middle of Fig.~\ref{fig:motivation}, the NF and RF both have a relatively clean background and clear stationary areas, which can directly feed the low-frequency information to the following network. However, \textbf{the high-frequency features such as edges and textures can not be directly obtained from the measurement and are neglected by previous studies for video SCI}. For the network structure, convolution-based backbone architectures have long dominated visual modeling in computer vision~\cite{lecun1998gradient,simonyan2014very,huang2017densely,he2016deep}. It is the same in the video SCI tasks, all previous SOTA learning-based algorithms are CNN-based. Although CNN has many advantages, its receptive field is usually small and relies on deeper layers or larger convolution kernels, which is not conducive to capture global features such as contour features and texture features which are also the high-frequency features. 
By contrast, Transformer can well capture long-distance dependencies and global inter-dependence between different regions, yet few researches of applying Transformer to video SCI are carried out. 

To sum up, previous learning-based frameworks mainly suffer from following two problems: 1) High-frequency information is not taken into consideration. 2) Compared to Transformer, CNNs are weak in capturing global features, some of which are also high-frequency features like contour features and texture features. Due to the mutual influence of these two aspects in the reconstruction process, the fidelity of the high-frequency details is compromised.


\noindent{\bf {Contributions}:} Towards this end, hereby, we propose a {\em Transformer enabled deep unfolding framework} for video SCI and we further introduce {\em uncertainty estimation to take high-frequency information as regularized prior under the unfolding framework into consideration} for better reconstruction. Our contributions can be summarized as follows:
\begin{itemize}
\setlength{\itemsep}{-2pt}
    \item[1)] We propose a novel video {\bf Convolution-Transformer module}, dubbed CTM, for video SCI that can well capture local and global spatial-temporal interactions which is composed of 3D CNN, 3D scalable {\em blocked dense and dilated sparse attention}. Note that the attention modules take both local and global information into consideration with  only  a linear complexity.
    \item[2)] Unlike previous studies that only consider the low-frequency information such as the information of stationary areas or backgrounds~\cite{cheng2020birnat,Wu_2021_ICCV,cheng2021memory}, we first bring {\bf high-frequency information as regularized prior under the unfolding framework in video SCI for focusing on areas with high reconstruction uncertainty and improving the fidelity of reconstruction}, which is achieved by the variance estimation characterizing the {\em uncertainty} on a pixel-by-pixel basis. 
    \item[3)] We first introduce Transformer for video SCI reconstruction. Both real and simulation experiments demonstrate that \textbf{our proposed framework outperform previous SOTA algorithms with a large margin of PSNR over 1.2dB}.
\end{itemize}

\section{Related Work}
\noindent{\bf {Snapshot compressive imaging}:} In terms of hardware, except capturing high-speed video frames~\cite{hitomi2011video,gao2014single,Llull13_OE_CACTI}, SCI has demonstrated promising results on spectral\cite{Gehm07_DDCASSI,zheng2021deep,lin2014spatial}, spectral-temporal\cite{tsai2015spectral}, polarization\cite{tsai2015spatial}, and coherent diffraction imaging\cite{Chen:22}, etc. The underlying principle of these systems is to modulate the high dimensional signals and capture the measurement compressively.

From the software perspective, the reconstruction algorithms can be broadly divided into two categories, \ie, optimization-based and learning-based algorithms. Optimization-based methods utilize various priors~\cite{yuan2016generalized,zhao2016video,liu2018rank,Yang14GMMonline,Yang14GMM} during reconstruction. However, the inference time is limited by the iterative solution process. As the development of deep learning, learning-based algorithms achieve impressive success in solving inverse problems. Different network backbones, such as CNN\cite{qiao2020deep,cheng2021memory} and recurrent neural network (RNN)\cite{cheng2020birnat} have been employed for video SCI reconstruction. Though these learning-based methods can achieve more decent results, their reconstruction process lacks interpretability. Combining the merits of above both kind of methods, DUNs\cite{Wu_2021_ICCV,li2020end,9428320} have been developed. However, {\em pixels with high reconstruction variance} have not attracted enough attention.

\noindent{\bf {Uncertainty}:} Uncertainty has been widely studied to help solve the reliability assessment, regression, risk-based decision making problems\cite{der2009aleatory,faber2005treatment,pate1996uncertainties,goldberg1997regression,wright1999bayesian}. Recently, uncertainty has been introduced into deep learning to improve the robustness and performance of deep neural networks for computer vision tasks such as semantic segmentation\cite{isobe2017deep,kendall2015bayesian}, image classification\cite{gu2015active}, object detection\cite{choi2019gaussian}, etc. For uncertainties in deep learning, they can be roughly classified into model uncertainty capturing the noise of the network's parameters, and data uncertainty referring to the noise inherent in the training data. Ning \etal investigated the data uncertainty with estimated mean and variance in low-level vision task such as super-resolution\cite{ning2021uncertainty}, which focuses on the areas with higher variance and achieved better result. \textbf{In image or video restoration tasks, high-frequency information is hard to be reconstructed \cite{Xu_2020_CVPR} due to the corresponding high reconstruction uncertainty. } To introduce the high-frequency information from the measurement into the network during training process, we first estimate the uncertainty and extract the feature maps of high-frequency information which are finally fed into the unfolding framework as regularized prior. 


\noindent{\bf {Transformer for vision}:} 
Recently, Transformer has achieved impressive success in the field of natural language processing due to the powerful self-attention mechanism, which inspires numerous researchers to introduce attention mechanism into vision. Many works\cite{wang2018non,yin2020disentangled} provide a complementary component (Self-attention/Transformers) to CNNs for modeling long range dependency. Vision Transformer (ViT)\cite{dosovitskiy2020image} and its follow-ups\cite{han2021transformer,touvron2021training,wang2021pyramid,yuan2021tokens,tu2022maxvit} start the trend of that backbone architectures for computer vision shift from CNNs to Transformers. Swin Transformer\cite{liu2021swin} is a typical representative and the key design is its shift of the window partition between consecutive self-attention layers, which enables it to serve as a general backbone for various tasks. 
Video Swin Transformer\cite{liu2022video} extends the scope of local attention computation from only the spatial domain to the spatiotemporal domain through spatiotemporal adaptation of Swin Transformer. In this paper, our proposed CTM takes both spatiotemporal globally and locally into account by {\bf integrating Transformer and 3D-CNN}, and outperforms all previous SOTA methods.

\section{Review the Forward Model of Video SCI} 
The top-left of Fig.~\ref{fig:Network} depicts the principle of video SCI, where multiple high-speed frames $\Xmat\in\mathbb{R}^{W\times H\times T}$ are modulated by different masks $\Mmat\in\mathbb{R}^{W\times H\times T}$ and then the measurement $\Ymat\in\mathbb{R}^{W\times H}$ is captured by a 2D camera, where $W$, $H$, and $T$ denote the width, height, and the number of frames, respectively. The 2D measurement is
\begin{align}
\label{eq:forward}
\textstyle \Ymat=\sum_{t=1}^{T}\Xmat_t\odot \Mmat_t+\Nmat,
\end{align}
where $\Nmat\in\mathbb{R}^\mathit{WH}$ denotes the measurement noise and $\odot$ represents the Hadamard (element-wise) multiplication. Eq.~\eqref{eq:forward} can be rewritten as the following linear from:
\begin{equation}
\label{eq:linear}
  \yv= \Phimat \xv+ \nv,
\end{equation}
where $\xv=\mathtt{vec}(\Xmat')\in\mathbb{R}^\mathit{WHT}$, $\yv=\mathtt{vec}({\Ymat})\in\mathbb{R}^\mathit{WH}$, and $\nv=\mathtt{vec}({\Nmat})\in\mathbb{R}^\mathit{WH}$. $\mathtt{vec}()$ here denotes vectorization. The sensing matrix $\Phimat\in\mathbb{R}^{\mathit{WH}\times \mathit{WHT}}$ can be expressed as
\begin{equation}
\label{eq:phi}
 \textstyle  \Phimat = [\mathtt{Diag}(\mathtt{vec}(\Mmat_1)),\dots,\mathtt{Diag}(\mathtt{vec}(\Mmat_t))].
\end{equation}
$\mathtt{Diag}()$ here means diagonalizing the vector. Note that $\Phimat$ is a very sparse matrix and the reconstruction error is bounded when $T>1$ \cite{jalali2019snapshot}.

\section{Proposed Method}

\noindent{\bf DUN Framework:}
SCI reconstruction is an ill-posed problem which can be modeled as:
\begin{equation}
\label{eq:oriproblem}
  \xv = \textstyle {\arg\min}_{\xv} \|\yv - \Phimat \xv\|_2^2 + \lambda \psi(\xv), 
\end{equation}
where $\psi(\xv)$ denotes the regularization term to confine the solutions, $\lambda$ balances the two terms. Here we unfold the iterations utilizing the framework of generalized alternating projection (GAP)~\cite{Liao14GAP}, which solves:
\begin{equation}
\label{eq:gapproblem}
  \{\hat{\xv},\hat{\vv}\} = \textstyle {\arg\min}_{\xv} \|\xv - \vv\|_2^2 + \lambda \psi(\vv), ~~s.t.~~ \yv= \Phimat \xv.
\end{equation}
The solution can be derived by the following two steps:
\begin{itemize}[leftmargin=*]
\setlength{\itemsep}{0pt}
\setlength{\parsep}{0pt}
\setlength{\parskip}{0pt}
    \item Given $\vv$, $\xv$ is updated by the following projection:
    \begin{equation}
    \xv^{(j)}=\vv^{(j-1)}+\Phimat\tsp(\Phimat\Phimat\tsp)^{-1}(\yv-\Phimat\vv^{(j-1)}).   
    \label{eq:pnpGAP1}
    \end{equation}
    Recall Eq.~\eqref{eq:phi}, we have $\Phimat\Phimat\tsp = \mathtt{Diag}(R_1,\dots,R_{WH})$ is a diagonal matrix where $R_i=\sum_{t=1}^{T}\Mmat_{t,i}^2$, $\forall i=1,\dots,WH$. Thus Eq.~\eqref{eq:pnpGAP1} can be efficiently solved.
    \item Given $\xv$, $\vv$ is achieved by:
    \begin{equation}
     \vv^{(j)}=\Theta([\xv^{(j)},\Gamma]), \label{eq:pnpGAP2} 
    \end{equation}
    where $\vv^{(j)}$ denotes the $j$-th phase's estimate of the target signal, $[\cdot]$ denotes the concatenation, $\Gamma$ represents other inputs of different phases, and $\Theta$ symbolizes the proposed prior module in each phase. To balance the trade-off between reconstruction performance and model size, we only utilize 3 phases which will be disscused in Sec.\ref{sec:ablation}. Unlike conventional optimization-based algorithms utilizing various denoisers, in most unfolding-based algorithms deep networks are used to learn a more appropriate prior to constrain the signal domain. \textbf{Differently, we do not just let the network to learn a prior and we further introduce a regularized prior input into our unfolding framework by uncertainty estimation which focuses on the pixels with higher reconstruction uncertainty.} More network structure details in the following sections.
\end{itemize}

\noindent{\bf Uncertainty Estimation for SCI:}
As mentioned above, uncertainty could be roughly classified into model uncertainty capturing the noise of the network's parameters, and data uncertainty referring to the noise inherent in given training data\cite{kendall2017uncertainties}. 
We investigate the data uncertainty estimation for SCI. Let $f(\cdot)$ denotes the reconstruction algorithm, the data uncertainty can be formed as an additive term $\sigma$. In this way, the observation model can be formulated as:
    \begin{equation}
     \xv=\fv(\yv)+{\epsilonv}{\sigma}, \label{eq:observation} 
    \end{equation}
where $\epsilonv\sim{\cal N}(0,1)$. 
We assume a Gaussian distribution to characterize the likelihood function: 
    \begin{equation}
     \pv(\xv,\sigma|\yv)= \textstyle \frac{1}{\sqrt{2\piv}\sigma}\exp(-\frac{\|\xv-\fv(\yv)\|^2_2}{2{\sigma}^2})
, \label{eq:likelihood} 
    \end{equation}
the log-likelihood function is naturally represented as:
    \begin{equation}
    \ln \pv(\xv,\sigma|\yv)= \textstyle-\frac{\|\xv-\fv(\yv)\|^2_2}{2{\sigma}^2}-\frac{1}{2}\ln\sigma^2-\frac{1}{2}\ln2\piv.
    \label{eq:log likelihood} 
    \end{equation}
As shown in the top-right of Fig.~\ref{fig:Network}, we learn the target estimation (mean value, $f(y)$) and uncertainty (variance, $\sigma^2$) respectively by two decoding branches sharing the same encoder. Note that the network $f(\cdot)$ here has the same structure as the network in each phase except the additional decoding branch for uncertainty estimation, we will talk about this in the Sec.\ref{sec:ablation}.
For more stable training, we estimate the log variance $\beta=\ln\sigma^2$ rather than directly estimate $\sigma^2$ due to the high dynamic range. Maximizing the likelihood in Eq.\eqref{eq:log likelihood} is same as minimizing the following loss function for learning the uncertainty (variance) of SCI reconstruction:
\begin{equation}
    \mathcal{L}_{U}=\exp(-\beta)\|\xv-\fv(\yv)\|^2_2+\beta. \label{eq:uncertainty loss} 
\end{equation}
\begin{figure}
\centering
\includegraphics[width=0.9\columnwidth]{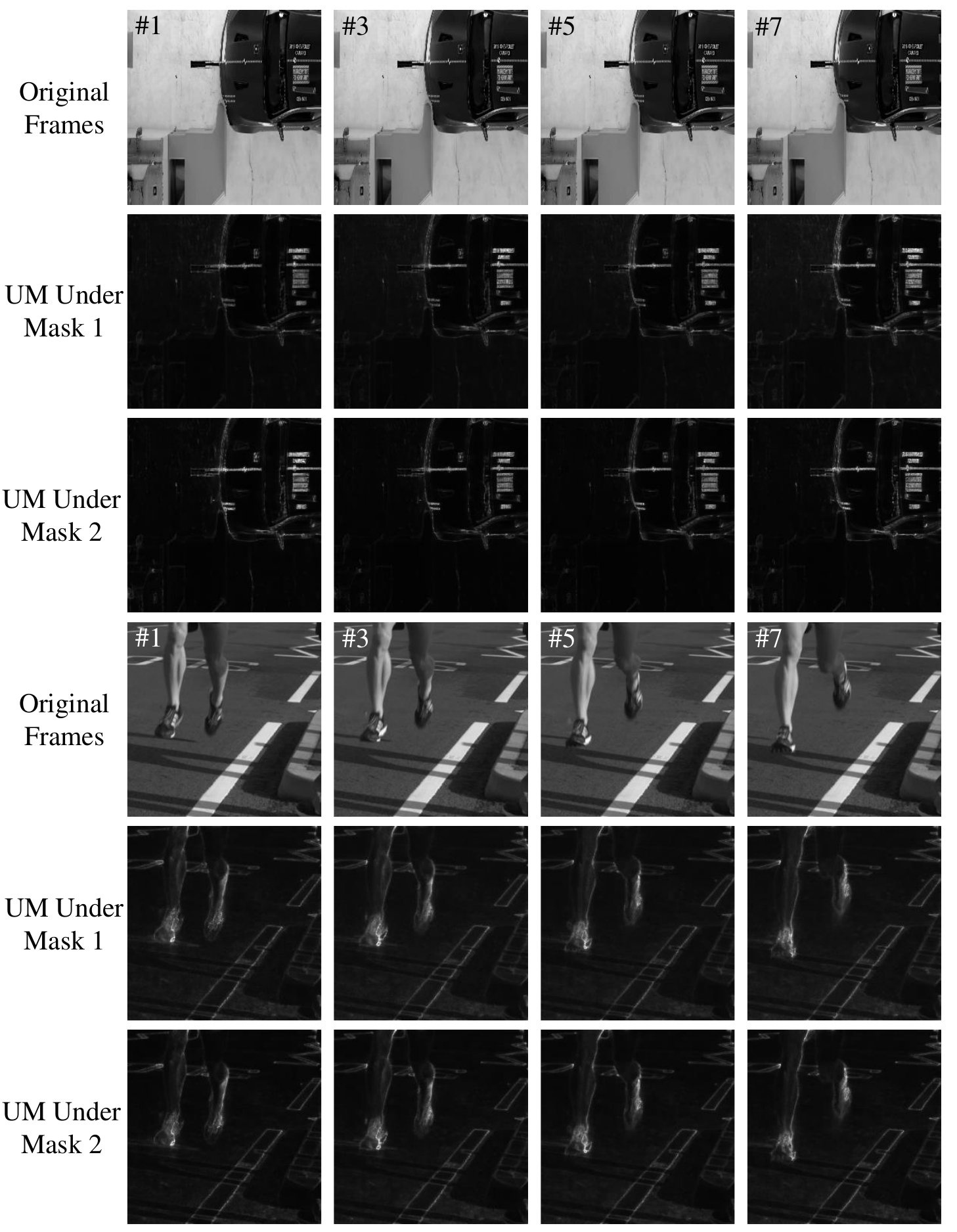}
\caption{{Visualization of the estimated uncertainty of two selected scenes under two different masks.}}
\vspace{-5mm}
\label{fig:uncertainty}
\end{figure}
The uncertainty estimation results is shown in Fig.~\ref{fig:uncertainty}. To visually highlight the pixel with high variance, we utilize thresholding method for the binarization processing in Fig.~\ref{fig:uncertainty}, and the threshold is the mean of the intensity. \textbf{We can observe that pixels with high variance are distributed around the high-frequency details, such as edges and textures.} In previous researches~\cite{cheng2020birnat,Wu_2021_ICCV,cheng2021memory}, Reference Frames (RF) are utilized in the initialization part for introducing the low-frequency information to improve the reconstruction performance. As shown in the top-left of Fig.~\ref{fig:Network}, the Normalized Measurement can achieve more visually clear background and stationary areas {\em but with blurry edges and textures.} Hence, at the initialization of each phase, we not only focus on the low-frequency information but also take the high-frequency details into consideration. The feature maps being fed into each phase are extracted from the estimated uncertainty map (UM) by three 3D-CNN blocks. The 3D-CNN blocks have the same structure but without sharing parameters.
Fig.~\ref{fig:uncertainty} also shows the uncertainty estimation module's adaptability to different masks. Under  different masks, the uncertainty estimation is unaffected.


\noindent{\bf Convolution-Transformer Mixture:}
Multi-head self-attention modules (MSA) are widely used in Transformers. Most of traditional MSAs of Transformers for video perform global spatial interactions by utilizing all tokens extracted from the whole feature map, which requires quadratic complexity. Compared to images, videos need to take the correlation of temporal dimension into consideration. Inspired by previous studies~\cite{liu2021swin,liu2022video,tu2022maxvit}, we propose a novel attention module. As shown in the bottom of Fig.~\ref{fig:Network}, CTM is composed of three sequential stacked parts, \ie, 3D blocked dense attention (BDA) for local interaction, 3D dilated sparse attention (DSA) for global interaction, and 3D-CNN based feature fusion (FF) module for further exploring spatiotemporal correlations. 

Let $\Xmat_f\in\mathbb{R}^{W\times H\times T\times C}$denote input feature map. In BDA, given a 3D window size of $P\times P\times M$, the input tokens are partitioned into  $\frac{W}{P}\times \frac{H}{P}\times \frac{T}{M}$ non-overlapping 3D windows. As shown in the bottom of Fig.~\ref{fig:Network}, given an input with the size of $8\times 8\times 8$ and the 3D window size $4\times 4\times 4$, we achieve 8 3D windows. And we conduct MSA on each window:
\begin{equation}
    {\rm MSA}(\Xmat_f)={\rm Softmax}(QK^T/\sqrt{d}+B)V, 
    \label{eq:MSA} 
\end{equation}
where $Q, K, V$ denotes the query, key, and value matrix respectively, the number of each head's channels $d=\frac{C}{N}$ and $N$ is the number of heads. $B$ represents the 3D relative position bias. compared to full self-attention (FSA), 
\begin{equation}
     \Omega(\rm FSA)=4WHTC^2+2(WHT)^2C. 
    \label{eq:FSA complexity} 
\end{equation}
BDA allows local spatiotemporal interactions with only a linear complexity,
\begin{equation*}
  \begin{aligned}
    \Omega(\rm BDA) &= \textstyle [4P^2MC^2+2(P^2M)^2C]\frac{WHT}{P^2M},\\
         &= 4WHTC^2+2WHTP^2MC.
  \end{aligned}
  \label{eq:BDA complexity} 
\end{equation*}
For practical applications of SCI, large-scale scenarios are very common. However, local-attention models do not adapt well to large scales\cite{dai2021coatnet,dosovitskiy2020image}. Inspired by \cite{zhang2018shufflenet}, we propose 3D DSA for global interaction. Unlike BDA where the input tokens are partitioned into non-overlapping 3D windows, in DSA, to keep the fixed group size of $S\times S\times B$, the tokens are selected from the sparse positions with the interval of $\frac{W}{S}\times \frac{H}{S}\times \frac{T}{B}$. As shown in the bottom of Fig.~\ref{fig:Network}, given an input with the size of $8\times 8\times 8$ and the interval size of $2\times 2\times 2$, we achieve 8 groups with the size of $4\times 4\times 4$ and employ MSA as well. Note that the complexity of DSA for global interaction is also linear,
\begin{equation*}
  \begin{aligned}
    \Omega(\rm DSA) &= \textstyle [4S^2BC^2+2(S^2B)^2C]\frac{WHT}{S^2B} \\
         &= 4WHTC^2+2WHTS^2BC.
  \end{aligned}
  \label{eq:DSA complexity} 
\end{equation*}
Recall video Swin\cite{liu2022video} where the mechanism of 3D shifted windows is employed to bridge the connections across different windows, our proposed 3D local and global attention achieves this in a more implementation friendly way and is scalable. 

We propose an initialization feature extraction block at the beginning of each phase to increase the generalization and trainability of the network. In each CTM block, to further explore the correlation of spatiotemporal dimensions, we plug FF into each CTM block. In FF, the feature map is first divided into two parts according to the channels. Then the two parts with skip connection are respectively sent into two Resnet modules with the same structure but not sharing parameters. Finally the features are fused to keep the original dimensions. We utilize 3D-CNN for all the convolutional layers.

\noindent{\bf Training:} Prior to the training of uncertainty estimation, we first train the whole network without uncertainty estimation to ensure the convergence. Given the training pairs ${(y_i,x_i)}_{i=1}^{N}$, where $N$ is training data number (52000 cropped pairs used here), the mean square error (MSE) loss is selected as the loss function. After 20 epoch training, $\mathcal{L}_{U}$ loss function is utilized to estimate the uncertainty. The initial learning rate is $5e^{-5}$ for the first 10 epochs and decays to $1e^{-5}$ for the last 10 epochs. After the training of UM estimation, we fix all the parameters of uncertainty estimation network and train the proposed framework with the initialization of corresponding parameters from the same network modules, \ie, duplicating the corresponding parameters from the uncertainty estimation network to each phase, which will lead to faster convergence of training.

The network is trained on 2 NVIDIA A40 GPUs utilizing PyTorch~\cite{paszke2019pytorch}. Adam~\cite{kingma2014adam} is employed as the optimizer. Note that, for the training of uncertainty estimation, if we directly use $\mathcal{L}_{U}$ for training, the training is easy to diverge. Therefore, we used MSE loss for training at the first. $S\times S\times B$ and $P\times P\times M$ we set in the experiments are the same, \ie, $7\times 7\times 2$. The setting of the spatial parameters, \ie, $P$ and $S$, follows Swin Transformer\cite{liu2021swin}. And the chosen number of $B$ and $M$, \ie, two, echoes the two divided parts in the FF module.

\begin{table*}[!h]
\centering
	\resizebox{1\textwidth}{!}
	{
\begin{tabular}{
>{\columncolor[HTML]{FFFFFF}}c 
>{\columncolor[HTML]{FFFFFF}}c 
>{\columncolor[HTML]{FFFFFF}}c 
>{\columncolor[HTML]{FFFFFF}}c 
>{\columncolor[HTML]{FFFFFF}}c 
>{\columncolor[HTML]{FFFFFF}}c 
>{\columncolor[HTML]{FFFFFF}}c 
>{\columncolor[HTML]{FFFFFF}}c 
>{\columncolor[HTML]{FFFFFF}}c }
\hline
{\color[HTML]{000000} Dataset}        & {\color[HTML]{000000} Kobe}                 & {\color[HTML]{000000} Traffic}     & {\color[HTML]{000000} Runner}               & {\color[HTML]{000000} Drop}        & {\color[HTML]{000000} Aerial}               & {\color[HTML]{000000} Crash}                & {\color[HTML]{000000} Average}              & {\color[HTML]{000000} Running time} \\ \hline
{\color[HTML]{000000} GAP-TV \cite{yuan2016generalized}}         & {\color[HTML]{000000} 26.45 0.845}          & {\color[HTML]{000000} 20.90 0.715} & {\color[HTML]{000000} 28.48 0.899}          & {\color[HTML]{000000} 33.81 0.963} & {\color[HTML]{000000} 25.03 0.828}          & {\color[HTML]{000000} 24.82 0.838}          & {\color[HTML]{000000} 26.58 0.848}          & {\color[HTML]{000000} 4.2}          \\
{\color[HTML]{000000} E2E-CNN~\cite{qiao2020deep}}        & {\color[HTML]{000000} 27.79 0.807}          & {\color[HTML]{000000} 24.62 0.840} & {\color[HTML]{000000} 34.12 0.947}          & {\color[HTML]{000000} 36.56 0.949} & {\color[HTML]{000000} 27.18 0.869}          & {\color[HTML]{000000} 26.43 0.882}          & {\color[HTML]{000000} 29.45 0.882}          & {\color[HTML]{000000} 0.0312}       \\
{\color[HTML]{000000} DeSCI~\cite{liu2018rank}}          & {\color[HTML]{000000} 33.25 0.952}          & {\color[HTML]{000000} 28.72 0.925} & {\color[HTML]{000000} 38.76 0.969}          & {\color[HTML]{000000} 43.22 0.993} & {\color[HTML]{000000} 25.33 0.860}          & {\color[HTML]{000000} 27.04 0.909}          & {\color[HTML]{000000} 32.72 0.935}          & {\color[HTML]{000000} 6180}         \\
{\color[HTML]{000000} PnP-FFDNet \cite{yuan2020plug}}     & {\color[HTML]{000000} 30.47 0.926}          & {\color[HTML]{000000} 24.08 0.833} & {\color[HTML]{000000} 32.88 0.938}          & {\color[HTML]{000000} 40.87 0.988} & {\color[HTML]{000000} 24.02 0.814}           & {\color[HTML]{000000} 24.32 0.836}          & {\color[HTML]{000000} 29.44 0.889}          & {\color[HTML]{000000} 3.0}          \\
{\color[HTML]{000000} PnP-FastDVDNet~\cite{Yuan14CVPR}} & {\color[HTML]{000000} 32.73 0.946}          & {\color[HTML]{000000} 27.95 0.932} & {\color[HTML]{000000} 36.29 0.962}          & {\color[HTML]{000000} 41.82 0.989} & {\color[HTML]{000000} 27.98 0.897}          & {\color[HTML]{000000} 27.32 0.925}          & {\color[HTML]{000000} 32.35 0.942}          & {\color[HTML]{000000} 18}           \\
{\color[HTML]{000000} BIRNAT~\cite{cheng2020birnat}}         & {\color[HTML]{000000} 32.71 0.950}          & {\color[HTML]{000000} 29.33 0.942} & {\color[HTML]{000000} 38.70 0.976}          & {\color[HTML]{000000} 42.28 0.992} & {\color[HTML]{000000} 28.99 0.927}          & {\color[HTML]{000000} 27.84 0.927}          & {\color[HTML]{000000} 33.31 0.951}          & {\color[HTML]{000000} 0.16}         \\
{\color[HTML]{000000} GAP-Unet-S12~\cite{meng2020gap}}   & {\color[HTML]{000000} 32.09 0.944}          & {\color[HTML]{000000} 28.19 0.929} & {\color[HTML]{000000} 38.12 0.975}          & {\color[HTML]{000000} 42.02 0.992} & {\color[HTML]{000000} 28.88 0.914}          & {\color[HTML]{000000} 27.83 0.931}          & {\color[HTML]{000000} 32.86 0.947}          & {\color[HTML]{000000} 0.0072}       \\
{\color[HTML]{000000} MetaSCI~\cite{wang2021metasci}}        & {\color[HTML]{000000} 30.12 0.907}          & {\color[HTML]{000000} 26.95 0.888} & {\color[HTML]{000000} 37.02 0.967}          & {\color[HTML]{000000} 40.61 0.985} & {\color[HTML]{000000} 28.31 0.904}          & {\color[HTML]{000000} 27.33 0.906}          & {\color[HTML]{000000} 31.72 0.926}          & {\color[HTML]{000000} 0.025}        \\
{\color[HTML]{000000} RevSCI~\cite{cheng2021memory}}         & {\color[HTML]{000000} 33.72 0.957}          & {\color[HTML]{000000} 30.02 0.949} & {\color[HTML]{000000} 39.40 0.977}          & {\color[HTML]{000000} 42.93 0.992} & {\color[HTML]{000000} 29.35 0.924}          & {\color[HTML]{000000} 28.12 0.937}          & {\color[HTML]{000000} 33.92 0.956}          & {\color[HTML]{000000} 0.19}         \\
{\color[HTML]{000000} DUN-3DUnet~\cite{Wu_2021_ICCV}}     & {\color[HTML]{000000} {35.00} {0.969}}          & {\color[HTML]{000000} {31.76} {0.966}} & {\color[HTML]{000000} {40.90} {0.983}}          & {\color[HTML]{000000} {44.46} {0.994}} & {\color[HTML]{000000} {30.46} {0.943}}          & {\color[HTML]{000000} {29.35} {0.955}}          & {\color[HTML]{000000} {35.32} {0.968}}          & {\color[HTML]{000000} 1.35}         \\ \hline
{\color[HTML]{000000} Ours }           & {\color[HTML]{000000} \underline{35.77} \underline{0.984}} & {\color[HTML]{000000} \underline{32.40} \underline{0.979}} & {\color[HTML]{000000} \underline{41.82} \underline{0.993}} & {\color[HTML]{000000} \underline{45.25} \underline{0.996}} & {\color[HTML]{000000} \underline{31.41} \underline{0.968}} & {\color[HTML]{000000} \underline{31.08} \underline{0.978}} & {\color[HTML]{000000} \underline{36.29} \underline{0.983}} & {\color[HTML]{000000} 1.26}  \\ 
{\color[HTML]{000000} Ours-with-Uncertainty }           & {\color[HTML]{000000} \textbf{35.97 0.986}} & {\color[HTML]{000000} \textbf{32.59} \textbf{0.981}} & {\color[HTML]{000000} \textbf{42.10 0.995}} & {\color[HTML]{000000} \textbf{45.49} \textbf{0.998}} & {\color[HTML]{000000} \textbf{31.64 0.970}} & {\color[HTML]{000000} \textbf{31.33 0.980}} & {\color[HTML]{000000} \textbf{36.52 0.985}} & {\color[HTML]{000000} 1.58}  \\ \hline
\end{tabular}}
\caption{ { The quantitative comparison of different algorithms. The average results of PSNR in dB (left entry), SSIM (right entry) and running time per measurement in seconds. Note that GAP-TV and DeSCI are running on CPU while others are on GPU. The best results are bold, and the second best are \underline{underlined}. Full results are in SM.}}
\label{table:gray}
\end{table*}

\begin{table*}

\centering
	\resizebox{1\textwidth}{!}
	{
\begin{tabular}{ccccccccc}
\hline
 Size                       & Algorithm  & Beauty               & Bosphorus            & HoneyBee             & Jockey               & ShakeNDry            & Average              & Running Time \\ \hline
 \multirow{4}{*}{512x512}   & GAP-TV \cite{yuan2016generalized}     & 32.13 0.857          & 29.18 0.934          & 31.40 0.887          & 31.01 0.940          & 32.52 0.882          & 31.25 0.900          & 44.67     \\
                           & PnP-FFDNet \cite{yuan2020plug} & 30.70 0.855          & 35.36 {0.952}          & 31.94 0.872          & 34.88 0.955          & 30.72 0.875          & 32.72 0.902          & 14.22     \\
                          & MetaSCI~\cite{wang2021metasci}    & {35.10} {0.901}          & {38.37} 0.950          & {34.27} {0.913}          & {36.45} {0.962}          & {33.16} {0.901}          & {35.47} {0.925}          & 0.12      \\
                           & Ours       & \underline{41.22} \underline{ 0.983} & \underline{42.39 } \underline{0.990} & \underline{43.63} \underline{0.990} & \underline{41.81} \underline{0.988} & \underline{37.09 } \underline{0.966} & \underline{41.23 } \underline{0.983} & 4.97      \\
                           & Ours-with-Uncertainty       & \textbf{41.36 0.984} & \textbf{42.59 0.990} & \textbf{43.71 0.991} & \textbf{42.10 0.989} & \textbf{37.40 0.966} & \textbf{41.41 0.984} & 6.32      \\\hline
Size                       & Algorithm  & Beauty               & Jockey               & ShakeNDry            & ReadyGo              & YachtRide            & Average              & Test Time \\ \hline
\multirow{4}{*}{1024x1024} & GAP-TV \cite{yuan2016generalized}     & 33.59 0.852          & 33,27 0.971          & 33.86 0.913          & 27.49 0.948          & 24.39 0.937          & 30.52 0.924          & 178.11    \\
                           & PnP-FFDNet \cite{yuan2020plug} & 32.36 0.857          & 35.25 0.976          & 32.21 0.902          & 31.87 0.965          & 30.77 {0.967}          & 32.49 0.933          & 52.47     \\
                           & MetaSCI~\cite{wang2021metasci}    & {35.23} {0.929}          & {37.15} {0.978}          & {36.06} {0.939}          & {33,34} {0.973}          & {32.68} 0.955          & {34.89} 0.955          & 0.59      \\
                           & Ours       & \underline{40.11} \underline{0.978} & \underline{42.28} \underline{0.988} & \underline{38.95} \underline{0.978} & \underline{40.39} \underline{0.989} & \underline{37.76} \underline{0.982} & \underline{39.90} \underline{0.983} & 23.76     \\ 
                           & Ours-with-Uncertainty       & \textbf{40.40 0.979} & \textbf{42.46 0.990} & \textbf{39.22 0.979} & \textbf{40.60 0.989} & \textbf{37.96 0.983} & \textbf{40.13 0.984} & 31.78     \\\hline
Size                       & Algorithm  & City                 & Kids                 & Lips                 & RaceNight                & RiverBank            & Average              & Test Time \\ \hline
\multirow{4}{*}{2048x2048} & GAP-TV \cite{yuan2016generalized}     & 21.27 0.902          & 26.05 0.956          & 26.46 0.890          & 26.81 0.875          & 27.74 0.848          & 25.67 0.894          & 764.75    \\
                           & PnP-FFDNet \cite{yuan2020plug} & 29.31 0.926          & 30.01 {0.966}          & 27.99 {0.902}          & 31.18 0.891          & 30.38 0.888          & 29.77 0.915          & 205.62    \\
                           & MetaSCI~\cite{wang2021metasci}    & {32.63} {0.930}          & {32.31} 0.965          & {30.90} 0.895         & {33.86} {0.893}          & {32.77} {0.902}          & {32.49} {0.917}          & 2.38      \\
                           & Ours       & \underline{40.31} \underline{0.981} & \underline{40.22} \underline{0.984} & \underline{35.26} \underline{0.933} & \underline{36.36} \underline{0.924} & \underline{36.87} \underline{0.970} & \underline{37.81} \underline{0.964} & 95.06      \\ 
                           & Ours-with-Uncertainty       & \textbf{40.54 0.983} & \textbf{40.45 0.985} & \textbf{35.49 0.934} & \textbf{36.59 0.956} & \textbf{37.10 0.971} & \textbf{38.04 0.966} & 120.09     \\\hline
\end{tabular}}
\caption{Large-scale results (CR: 8): quantitative comparison of existing algorithms that can be applied to large-scale data. The best results are in bold, and the second best results are \underline{underlined}. PSNR and SSIM are selected as the evaluation metrics.}
\vspace{-5mm}
\label{table:scale result}
\end{table*}
\begin{figure}[!htbp]
\centering
\includegraphics [width=1\columnwidth]{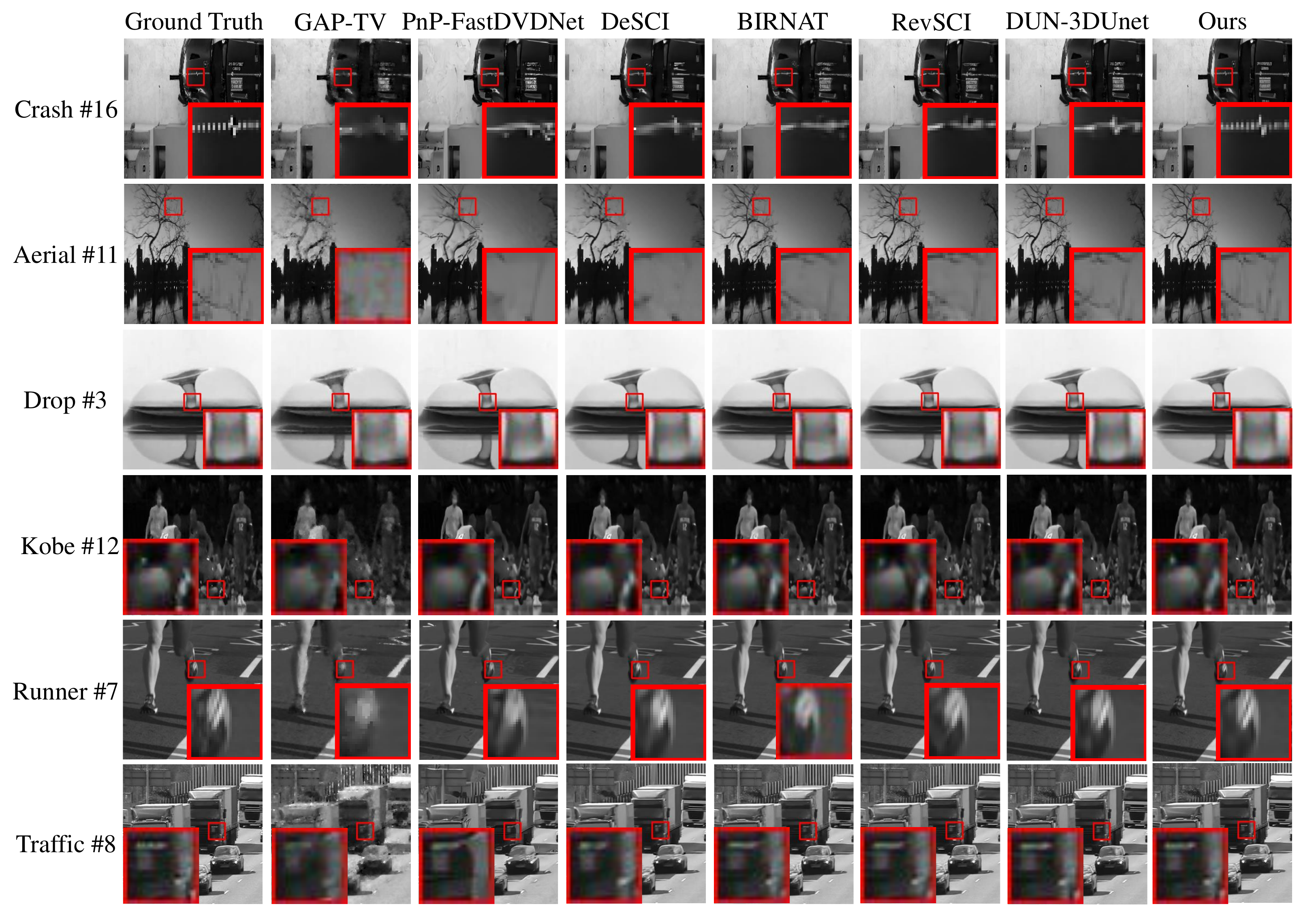}
\caption{Selected multiple reconstruction frames of simulated benchmark dataset.}
\vspace{-8mm}
\label{fig:simulation}
\end{figure}

\section{Experiments}
\noindent{\textbf{Dataset}:}
We choose \textbf{DAVIS 2017}~\cite{pont20172017} as our training dataset following previous studies. It contains 90 scenes with two resolutions: 480P and 1080P. We conduct data augmentation by random cropping, rotation and flip.

\begin{figure}
\centering
\includegraphics[width=1\columnwidth]{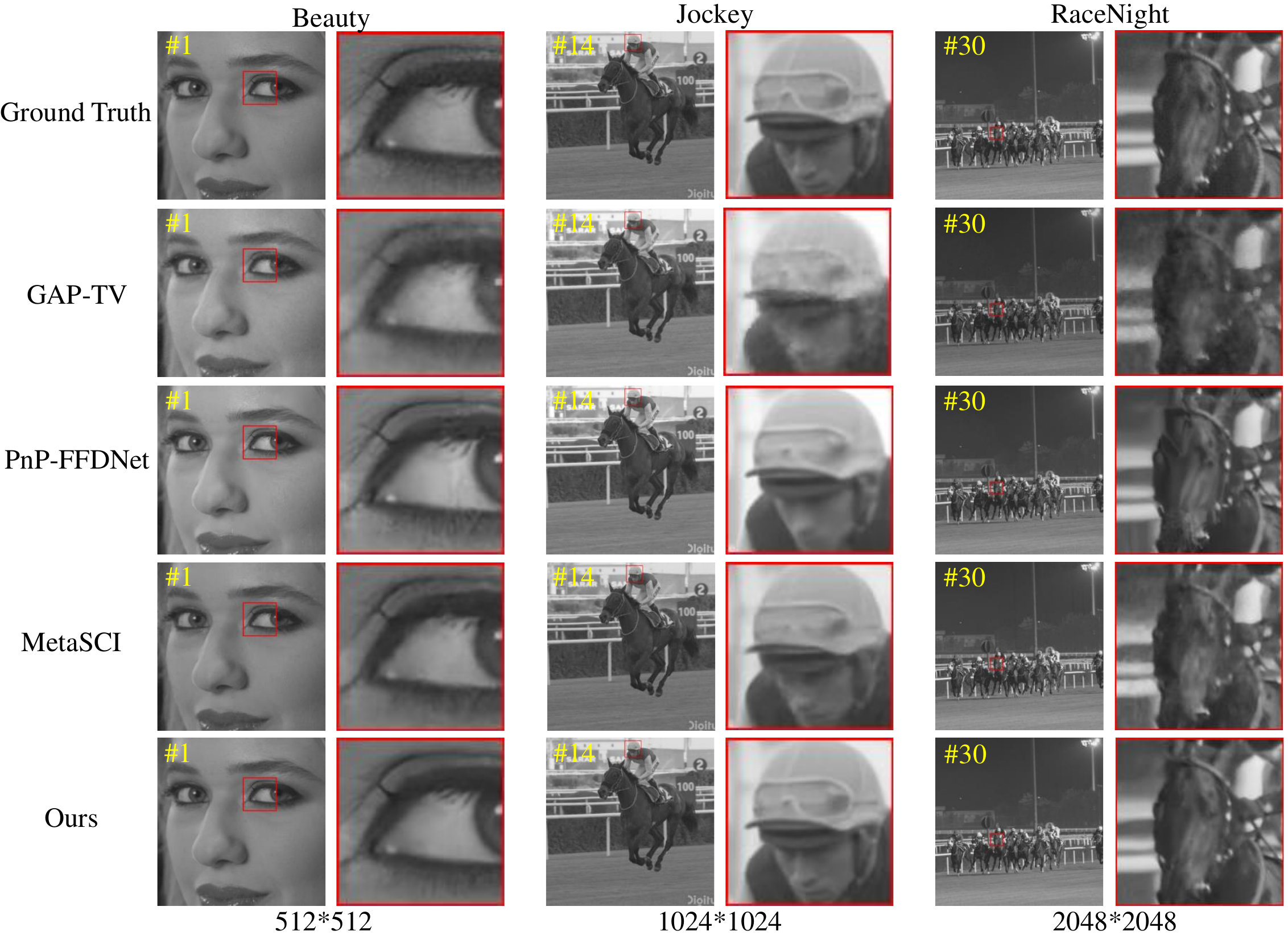}
\caption{{Selected reconstruction frames of different scales. Zoom in for better view.}}
\label{fig:large_scale_detail}
\vspace{-8mm}
\end{figure}

\subsection{Benchmark Simulation of SCI}
The testing synthetic datasets of Benchmark follow previous study~\cite{liu2018rank} including \textbf{Kobe, Traffic, Runner, Drop, Crash} and \textbf{Aerial} with the size of $256\times256\times8$.
We compare our model with previous SOTA algorithms, \ie, GAP-TV~\cite{yuan2016generalized}, PnP-FFDnet~\cite{yuan2020plug}, PnP-FastDVDnet~\cite{Yuan14CVPR}, DeSCI~\cite{liu2018rank},  E2E-CNN~\cite{qiao2020deep}, GAP-Unet-S12~\cite{meng2020gap}, BIRNAT~\cite{cheng2020birnat}, MetaSCI~\cite{wang2021metasci}, RevSCI~\cite{cheng2021memory} and DUN-3DUnet~\cite{Wu_2021_ICCV}.  The quantitative comparison is summarized in Tab.~\ref{table:gray}. Both PSNR and structured similarity (SSIM)~\cite{wang2004image} are selected to evaluate the reconstruction quality. It can be observed that our method substantially outperforms (by a large margin of nearly 1.2dB in PSNR) all previous SOTA algorithms. Selected reconstructed frames are shown in Fig.~\ref{fig:simulation}. As we can see, the optimization-based algorithms, such as GAP-TV and PnP, usually lead to over-smooth (Crash,Kobe,Runner, and Traffic) artifacts. DeSCI is with poor restoration of the irregular textures (Aerial). When the object is with large motion, other learning-based methods do not work well. Obviously, our proposed method achieves much better visually results on the areas with high uncertainty (variance), such as the edges, textures, and other high-frequency details. The inference time is on par with previous SOTA DUN-3DUnet.

\begin{table}[!h]
\centering
	\resizebox{0.47\textwidth}{!}
	{
\begin{tabular}{
>{\columncolor[HTML]{FFFFFF}}c 
>{\columncolor[HTML]{FFFFFF}}c 
>{\columncolor[HTML]{FFFFFF}}c 
>{\columncolor[HTML]{FFFFFF}}c 
>{\columncolor[HTML]{FFFFFF}}c 
>{\columncolor[HTML]{FFFFFF}}c 
>{\columncolor[HTML]{FFFFFF}}c 
>{\columncolor[HTML]{FFFFFF}}c 
>{\columncolor[HTML]{FFFFFF}}c }
\hline
{\color[HTML]{000000} Evaluation
metrics}        & {\color[HTML]{000000} Trained mask}                 & {\color[HTML]{000000} New mask 1}     & {\color[HTML]{000000} New mask 2}                \\ \hline
{\color[HTML]{000000} PSNR SSIM}         & {\color[HTML]{000000} 36.52 0.985}          & {\color[HTML]{000000} 36.47 0.985} & {\color[HTML]{000000} 36.48 0.985}                         \\\hline
\end{tabular}}
\caption{Quantitative comparison with different masks.}
\vspace{-5mm}
\label{table:Adaptability}
\end{table}

\noindent{\textbf{Adaptability}:}
We test our uncertainty estimation module under different masks. As shown in Fig.~\ref{fig:uncertainty}, it can well adapt to different masks. We further test the adaptability of the reconstruction, the results are presented in Tab.~\ref{table:Adaptability}. Note that all the experiments are directly conducted without training with other masks, which is never achieved by previous learning-based methods. (Other methods' results are in SM.)

\subsection{Scalability of Transformer on Large-scale Data}
As mentioned in the preceding part of the paper, the ability to cope with large-scale data is crucial for reconstruction algorithms. Our proposed scalable Transform module (BDA and DSA) facilitates the practical applications of SCI. We test the proposed model on the large-scale benchmark dataset \cite{wang2021metasci}. The quantitative comparison is summarized in Tab.~\ref{table:scale result}. As we can see, few algorithms can be applied to large scale data due to GPU memory limit while training, our proposed method far exceeds (nearly 6dB in PSNR) all previous SOTA algorithms with competitive inference time, which verifies our proposed Transformer module is with enough scalability to large-scale data. Details of selected reconstruction frames of different scales are shown in Fig.~\ref{fig:large_scale_detail}. It can be observed that we can achieve much better visual performance especially in the details.

\subsection{Ablation Study \label{sec:ablation}}
\noindent{\bf {Effectiveness of modules}:}
To validate the effectiveness of each part of our proposed CTM module, we conduct ablation experiments on the benchmark dataset for each sub-modules, \ie, BDA, DSA and FF. To reduce the effects of uncontrollable factors on the experiments, the above ablation experiments are conducted without uncertainty estimation with quantitative result shown in Tab.~\ref{table:Ablation of CTM}, where $\checkmark$ denotes the corresponding components are preserved, $\times$ is on the contrary. As we can observe, each of the modules is essential for the whole framework. 

As described in the above, to test the efficiency of each module, we directly remove each part of the module separately. However, \textbf{we should not ignore the effect brought by the reduction of parameter count.} In order to measure the effectiveness of the Transformer module more accurately, we conduct experiments utilizing BDA to replace DSA and utilizing DSA to replace DSA respectively, which all maintain the same parameter count and FLOPs. Block attention mechanism' efficiency has been verified in many other computer vision tasks ~\cite{liu2021swin}. However when we use BDA to replace DSA, PSNR decreases by 0.33dB on the benchmark dataset ($256*256*8$). When we use DSA to replace BDA, PSNR decreases by 0.87 dB on the benchmark dataset ($256*256*8$). The results demonstrate that local attention plays a more important role, yet the combination of both local and global attention leads to higher performance.
We also test \textbf{different order of the sub-modules,} \ie, BDA, DSA and FF in CTM block. Because the blocks are sequentially arranged, the change of the order of the sub-modules does not affect the performance.

\begin{figure}
\centering
\includegraphics[width=\columnwidth]{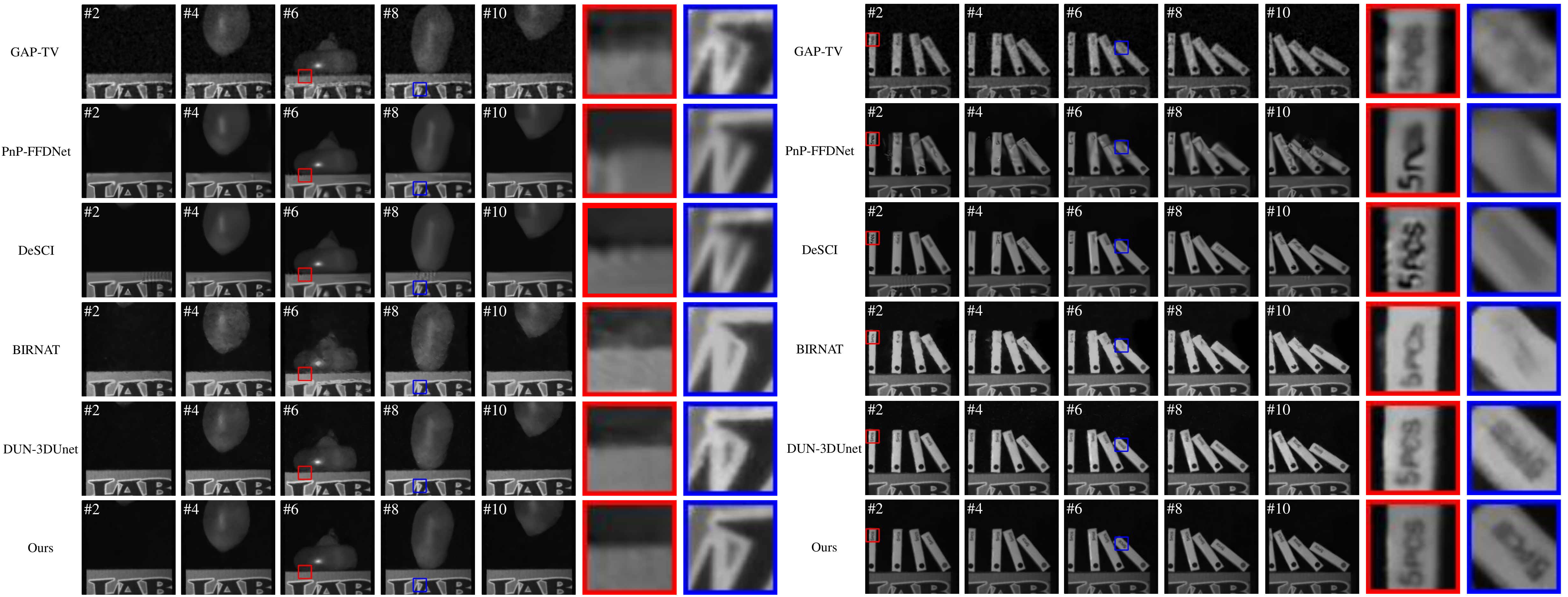}
\caption{{Selected reconstruction frames of real data \textbf{Water Balloon} and \textbf{Domino}. More results are in SM.}}
\vspace{-5mm}
\label{fig:waterballoon_domino}
\end{figure}

\begin{figure}
\centering
\includegraphics[width=\columnwidth]{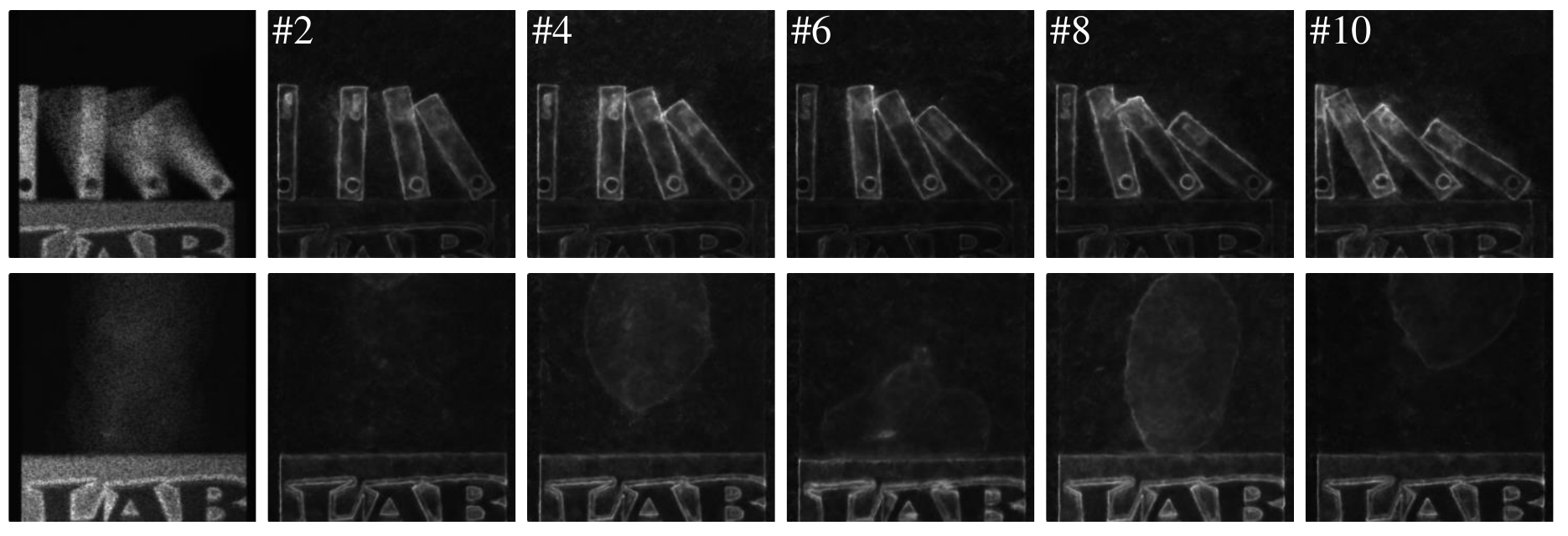}
\caption{{Selected estimated uncertainty map of real data \textbf{Water Balloon} and \textbf{Domino}.}}
\vspace{-5mm}
\label{fig:real_ncertainty}
\end{figure}

\begin{table}[!h]
\centering
	\resizebox{0.24\textwidth}{!}
	{
\begin{tabular}{
>{\columncolor[HTML]{FFFFFF}}c 
>{\columncolor[HTML]{FFFFFF}}c 
>{\columncolor[HTML]{FFFFFF}}c 
>{\columncolor[HTML]{FFFFFF}}c 
>{\columncolor[HTML]{FFFFFF}}c 
>{\columncolor[HTML]{FFFFFF}}c 
>{\columncolor[HTML]{FFFFFF}}c 
>{\columncolor[HTML]{FFFFFF}}c 
>{\columncolor[HTML]{FFFFFF}}c }
\hline
{\color[HTML]{000000} BDA}        & {\color[HTML]{000000} DSA}                 & {\color[HTML]{000000} FF}     & {\color[HTML]{000000} PSNR SSIM}                \\ \hline
{\color[HTML]{000000} $\times$}         & {\color[HTML]{000000} $\checkmark$}          & {\color[HTML]{000000} $\checkmark$} & {\color[HTML]{000000} 31.11 0.960}                         \\\hline
{\color[HTML]{000000} $\checkmark$}         & {\color[HTML]{000000} $\times$}          & {\color[HTML]{000000} $\checkmark$} & {\color[HTML]{000000} 31.04 0.955}                         \\\hline
{\color[HTML]{000000} $\checkmark$}         & {\color[HTML]{000000} $\checkmark$}          & {\color[HTML]{000000} $\times$} & {\color[HTML]{000000} 27.81 0.912}                         \\\hline
{\color[HTML]{000000} $\checkmark$}         & {\color[HTML]{000000} $\checkmark$}          & {\color[HTML]{000000} $\checkmark$} & {\color[HTML]{000000} 36.29 0.983}                         \\\hline
\end{tabular}}
\caption{Ablation study of CTM on benchmark dataset. The quantitative effects (PSNR in dB and SSIM) are shown.}
\vspace{-6mm}
\label{table:Ablation of CTM}
\end{table}

\noindent{\bf {Effectiveness of uncertainty estimation}:}
We also verify the effect of uncertainty estimation. Directly applying $\mathcal{L}_{U}$ loss for reconstruction brings a direct drop of nearly 1.1dB in PSNR. Recall Eq.~\eqref{eq:uncertainty loss}, the attention of pixels with high variance will be impaired by the division. Though paying less attention to pixels with high variance (uncertainty) helps promote performance in high level vision tasks\cite{kendall2017uncertainties,badrinarayanan2017segnet,chang2020data}, it does not work in low level vision tasks. As shown in Tab.~\ref{table:gray} and Tab.~\ref{table:scale result} , when the high-frequency information is not introduced, the PSNR and SSIM decline nearly 0.2-0.3dB and 0.002-0.003, respectively. Although we can further improve the reconstruction performance, the inference speed of the model is sacrificed. There is a trade-off between the benefits of reconstruction quality and the sacrifice of the inference speed. As mentioned above, we utilize the network of only one-phase instead of 
three-phases to estimate the uncertainty, which can reduce the inference time. We conducted experiments with the uncertainty map estimated by two different network, \ie, one-phase and three-phases networks, the reconstruction quality is almost the same. Obviously, one-phase uncertainty estimation has higher inference speed. Considering the memory cost, the phase number we chose is three in this paper. \textbf{The three-phases inference model with uncertainty estimation is basically with the same number of parameters as four-phases model.} Hence we test different phase numbers, \ie, 1, 2, 3 and 4, under our proposed framework without uncertainty estimation utilizing the same benchmark dataset ($256*256*8$). As shown in Tab.~\ref{table:phase number}, as the phase number increases, the reconstruction quality improvement is slowing down. Compared with three-phases, four-phases model only gain an increase of less than 0.1dB in PSNR, which is why we only use 3 phases and also illustrates the effectiveness of introducing uncertainty estimation.

\begin{table}[!h]
\centering
	\resizebox{0.47\textwidth}{!}
	{
\begin{tabular}{
>{\columncolor[HTML]{FFFFFF}}c 
>{\columncolor[HTML]{FFFFFF}}c 
>{\columncolor[HTML]{FFFFFF}}c 
>{\columncolor[HTML]{FFFFFF}}c 
>{\columncolor[HTML]{FFFFFF}}c 
>{\columncolor[HTML]{FFFFFF}}c 
>{\columncolor[HTML]{FFFFFF}}c 
>{\columncolor[HTML]{FFFFFF}}c 
>{\columncolor[HTML]{FFFFFF}}c }
\hline
{\color[HTML]{000000}Phase Number}        & {\color[HTML]{000000} One}                 & {\color[HTML]{000000} Two}     & {\color[HTML]{000000} Three}    & {\color[HTML]{000000} Four}      & {\color[HTML]{000000} Three with Uncertainty}       \\ \hline
{\color[HTML]{000000}  {PSNR SSIM}}         & {\color[HTML]{000000} 35.51 0.970}          & {\color[HTML]{000000} 36.12 0.981} & {\color[HTML]{000000} 36.29 0.983}  &{\color[HTML]{000000} 36.37 0.983}     &{\color[HTML]{000000} 36.52 0.985}                   \\\hline
\end{tabular}}
\caption{Reconstruction  with different phase numbers.}
\vspace{-7mm}
\label{table:phase number}
\end{table}



\subsection{Real Data Benchmark}
We test our model on the real data \textbf{Water Balloon} and \textbf{Domino} with the size of $512\times512\times10$~\cite{qiao2020deep}. Due to the uncontrollable noise during capturing, it is more challenging to reconstruct real measurements. Note that {\em we do not add any noise to the training data during the training with real masks, which demonstrates the generalization ability of our model to a certain extent}. The selected results are presented in Fig.~\ref{fig:waterballoon_domino}. In the areas with higher uncertainty (variance), such as edges and textures, our proposed method outperforms all existing algorithms, which is shown in the left part of Fig.~\ref{fig:waterballoon_domino}. Even when the water balloon collides with box, the edge of the box is still sharp in our reconstruction. 
Besides, falling dominoes are with higher speed, which further increases the difficulty of reconstruction. As we can observe in right part of Fig.~\ref{fig:waterballoon_domino}, all previous SOTA algorithms can not recover the legible letters except our proposed method. 
Our results are with sharper edges, more details, and cleaner background, which indicates our proposed method is more powerful in practical applications. The estimated uncertainty maps of real data are shown in Fig.~\ref{fig:real_ncertainty}, the edge and texture features can be directly obtained from the real measurement.

\begin{figure}
\centering
\includegraphics[width=1\columnwidth]{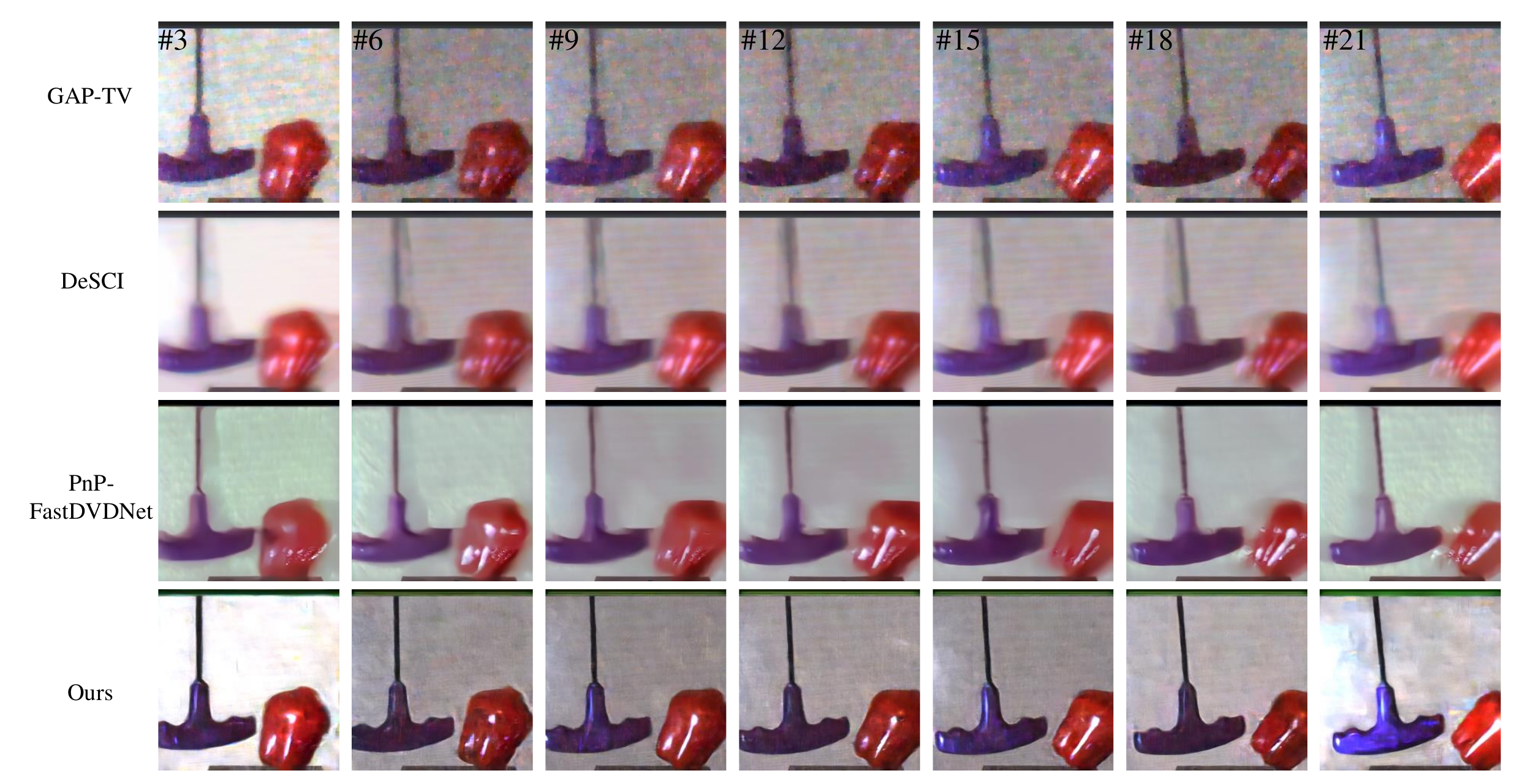}
\caption{{Comparison of selected reconstruction video frames of real color data \textbf{Hammer}.}}
\vspace{-6mm}
\label{fig:real hammer}
\end{figure}

We also test our model on real color dataset \textbf{Hammer} with the size of $512\times512\times22$. Few learning based algorithms conducted experiments on the color video SCI task. We compare our model with previous SOTA algorithms which are iteration-based. As we can see in Fig.~\ref{fig:real hammer}, GAP-TV has noisy results, DeSCI and PnP-FastDVDNet are blurry in the areas of background and edges, our results are cleaner and have sharp edges than other methods. The implementation details are in the SM.



\vspace{-2mm}
\section{Conclusions and Future Work}
\vspace{-2mm}
We have proposed a Transformer and 3D-CNN based network for video SCI reconstruction and introduced high-frequency information by uncertainty estimation. The design of the backbone with Transformer and 3D-CNN helps explore the correlation across the spatio-temporal dimensions. More importantly, our proposed method achieved SOTA results with a competitive inference time.

Although we have achieved the best results so far, the introduction of high-frequency information is time-inefficient and when the model is applied to large-scale data, 
the inference time is still long for real-time applications. In the future, we will reduce the parameters for high inference speed by knowledge distillation and employ the high-frequency information in a more time-efficient way. 
Besides video, our proposed framework can also be used in other inverse problems such as image compressive sensing and spectral compressive imaging.

\clearpage
\clearpage
\newpage

{\small
\bibliographystyle{ieee_fullname}
\bibliography{egbib}

\begin{thebibliography}{10}\itemsep=-1pt

\bibitem{badrinarayanan2017segnet}
Vijay Badrinarayanan, Alex Kendall, and Roberto Cipolla.
\newblock Segnet: A deep convolutional encoder-decoder architecture for image
  segmentation.
\newblock {\em IEEE transactions on pattern analysis and machine intelligence},
  39(12):2481--2495, 2017.

\bibitem{chang2020data}
Jie Chang, Zhonghao Lan, Changmao Cheng, and Yichen Wei.
\newblock Data uncertainty learning in face recognition.
\newblock In {\em Proceedings of the IEEE/CVF Conference on Computer Vision and
  Pattern Recognition}, pages 5710--5719, 2020.

\bibitem{Chen:22}
Ziyang Chen, Siming Zheng, Zhishen Tong, and Xin Yuan.
\newblock Physics-driven deep learning enables temporal compressive coherent
  diffraction imaging.
\newblock {\em Optica}, 9(6):677--680, Jun 2022.

\bibitem{cheng2021memory}
Ziheng Cheng, Bo Chen, Guanliang Liu, Hao Zhang, Ruiying Lu, Zhengjue Wang, and
  Xin Yuan.
\newblock Memory-efficient network for large-scale video compressive sensing.
\newblock In {\em Proceedings of the IEEE/CVF Conference on Computer Vision and
  Pattern Recognition}, pages 16246--16255, 2021.

\bibitem{cheng2020birnat}
Ziheng Cheng, Ruiying Lu, Zhengjue Wang, Hao Zhang, Bo Chen, Ziyi Meng, and Xin
  Yuan.
\newblock Birnat: Bidirectional recurrent neural networks with adversarial
  training for video snapshot compressive imaging.
\newblock In {\em European Conference on Computer Vision}, pages 258--275.
  Springer, 2020.

\bibitem{choi2019gaussian}
Jiwoong Choi, Dayoung Chun, Hyun Kim, and Hyuk-Jae Lee.
\newblock Gaussian yolov3: An accurate and fast object detector using
  localization uncertainty for autonomous driving.
\newblock In {\em Proceedings of the IEEE/CVF International Conference on
  Computer Vision}, pages 502--511, 2019.

\bibitem{dai2021coatnet}
Zihang Dai, Hanxiao Liu, Quoc~V Le, and Mingxing Tan.
\newblock Coatnet: Marrying convolution and attention for all data sizes.
\newblock {\em Advances in Neural Information Processing Systems},
  34:3965--3977, 2021.

\bibitem{der2009aleatory}
Armen Der~Kiureghian and Ove Ditlevsen.
\newblock Aleatory or epistemic? does it matter?
\newblock {\em Structural safety}, 31(2):105--112, 2009.

\bibitem{dosovitskiy2020image}
Alexey Dosovitskiy, Lucas Beyer, Alexander Kolesnikov, Dirk Weissenborn,
  Xiaohua Zhai, Thomas Unterthiner, Mostafa Dehghani, Matthias Minderer, Georg
  Heigold, Sylvain Gelly, et~al.
\newblock An image is worth 16x16 words: Transformers for image recognition at
  scale.
\newblock {\em arXiv preprint arXiv:2010.11929}, 2020.

\bibitem{faber2005treatment}
Michael~Havbro Faber.
\newblock On the treatment of uncertainties and probabilities in engineering
  decision analysis.
\newblock 2005.

\bibitem{gao2014single}
Liang Gao, Jinyang Liang, Chiye Li, and Lihong~V Wang.
\newblock Single-shot compressed ultrafast photography at one hundred billion
  frames per second.
\newblock {\em Nature}, 516(7529):74--77, 2014.

\bibitem{Gehm07_DDCASSI}
M.~E. Gehm, R. John, D.~J. Brady, R.~M. Willett, and T.~J. Schulz.
\newblock Single-shot compressive spectral imaging with a dual-disperser
  architecture.
\newblock {\em Opt. Express}, 15(21):14013--14027, Oct 2007.

\bibitem{goldberg1997regression}
Paul Goldberg, Christopher Williams, and Christopher Bishop.
\newblock Regression with input-dependent noise: A gaussian process treatment.
\newblock {\em Advances in neural information processing systems}, 10, 1997.

\bibitem{gu2015active}
Yingjie Gu, Zhong Jin, and Steve~C Chiu.
\newblock Active learning combining uncertainty and diversity for multi-class
  image classification.
\newblock {\em IET Computer Vision}, 9(3):400--407, 2015.

\bibitem{han2021transformer}
Kai Han, An Xiao, Enhua Wu, Jianyuan Guo, Chunjing Xu, and Yunhe Wang.
\newblock Transformer in transformer.
\newblock {\em Advances in Neural Information Processing Systems},
  34:15908--15919, 2021.

\bibitem{he2016deep}
Kaiming He, Xiangyu Zhang, Shaoqing Ren, and Jian Sun.
\newblock Deep residual learning for image recognition.
\newblock In {\em Proceedings of the IEEE conference on computer vision and
  pattern recognition}, pages 770--778, 2016.

\bibitem{hitomi2011video}
Yasunobu Hitomi, Jinwei Gu, Mohit Gupta, Tomoo Mitsunaga, and Shree~K Nayar.
\newblock Video from a single coded exposure photograph using a learned
  over-complete dictionary.
\newblock In {\em 2011 International Conference on Computer Vision}, pages
  287--294. IEEE, 2011.

\bibitem{huang2017densely}
Gao Huang, Zhuang Liu, Laurens Van Der~Maaten, and Kilian~Q Weinberger.
\newblock Densely connected convolutional networks.
\newblock In {\em Proceedings of the IEEE conference on computer vision and
  pattern recognition}, pages 4700--4708, 2017.

\bibitem{iliadis2020deepbinarymask}
Michael Iliadis, Leonidas Spinoulas, and Aggelos~K Katsaggelos.
\newblock Deepbinarymask: Learning a binary mask for video compressive sensing.
\newblock {\em Digital Signal Processing}, 96:102591, 2020.

\bibitem{isobe2017deep}
Shuya Isobe and Shuichi Arai.
\newblock Deep convolutional encoder-decoder network with model uncertainty for
  semantic segmentation.
\newblock In {\em 2017 IEEE International Conference on INnovations in
  Intelligent SysTems and Applications (INISTA)}, pages 365--370. IEEE, 2017.

\bibitem{jalali2019snapshot}
Shirin Jalali and Xin Yuan.
\newblock Snapshot compressed sensing: Performance bounds and algorithms.
\newblock {\em IEEE Transactions on Information Theory}, 65(12):8005--8024,
  2019.

\bibitem{kendall2015bayesian}
Alex Kendall, Vijay Badrinarayanan, and Roberto Cipolla.
\newblock Bayesian segnet: Model uncertainty in deep convolutional
  encoder-decoder architectures for scene understanding.
\newblock {\em arXiv preprint arXiv:1511.02680}, 2015.

\bibitem{kendall2017uncertainties}
Alex Kendall and Yarin Gal.
\newblock What uncertainties do we need in bayesian deep learning for computer
  vision?
\newblock {\em Advances in neural information processing systems}, 30, 2017.

\bibitem{kingma2014adam}
Diederik~P Kingma and Jimmy Ba.
\newblock Adam: A method for stochastic optimization.
\newblock {\em arXiv preprint arXiv:1412.6980}, 2014.

\bibitem{lecun1998gradient}
Yann LeCun, L{\'e}on Bottou, Yoshua Bengio, and Patrick Haffner.
\newblock Gradient-based learning applied to document recognition.
\newblock {\em Proceedings of the IEEE}, 86(11):2278--2324, 1998.

\bibitem{li2020end}
Yuqi Li, Miao Qi, Rahul Gulve, Mian Wei, Roman Genov, Kiriakos~N Kutulakos, and
  Wolfgang Heidrich.
\newblock End-to-end video compressive sensing using anderson-accelerated
  unrolled networks.
\newblock In {\em 2020 IEEE International Conference on Computational
  Photography (ICCP)}, pages 1--12. IEEE, 2020.

\bibitem{Liao14GAP}
X. Liao, H. Li, and L. Carin.
\newblock Generalized alternating projection for weighted-$\ell_{2,1}$
  minimization with applications to model-based compressive sensing.
\newblock {\em SIAM Journal on Imaging Sciences}, 7(2):797--823, 2014.

\bibitem{lin2014spatial}
Xing Lin, Yebin Liu, Jiamin Wu, and Qionghai Dai.
\newblock Spatial-spectral encoded compressive hyperspectral imaging.
\newblock {\em ACM Transactions on Graphics (TOG)}, 33(6):1--11, 2014.

\bibitem{liu2018rank}
Yang Liu and etc.
\newblock Rank minimization for snapshot compressive imaging.
\newblock {\em IEEE TPAMI}.

\bibitem{liu2021swin}
Ze Liu, Lin, and etc.
\newblock Swin transformer: Hierarchical vision transformer using shifted
  windows.
\newblock In {\em ICCV}, 2021.

\bibitem{liu2022video}
Ze Liu, Jia Ning, Yue Cao, Yixuan Wei, Zheng Zhang, Stephen Lin, and Han Hu.
\newblock Video swin transformer.
\newblock In {\em Proceedings of the IEEE/CVF Conference on Computer Vision and
  Pattern Recognition}, pages 3202--3211, 2022.

\bibitem{Llull13_OE_CACTI}
Patrick Llull, Xuejun Liao, Xin Yuan, Jianbo Yang, David Kittle, Lawrence
  Carin, Guillermo Sapiro, and David~J. Brady.
\newblock Coded aperture compressive temporal imaging.
\newblock {\em Opt. Express}, 21(9):10526--10545, May 2013.

\bibitem{meng2020gap}
Ziyi Meng, Shirin Jalali, and Xin Yuan.
\newblock Gap-net for snapshot compressive imaging.
\newblock {\em arXiv preprint arXiv:2012.08364}, 2020.

\bibitem{ning2021uncertainty}
Qian Ning, Weisheng Dong, Xin Li, Jinjian Wu, and Guangming Shi.
\newblock Uncertainty-driven loss for single image super-resolution.
\newblock {\em Advances in Neural Information Processing Systems},
  34:16398--16409, 2021.

\bibitem{paszke2019pytorch}
Adam Paszke, Sam Gross, Francisco Massa, Adam Lerer, James Bradbury, Gregory
  Chanan, Trevor Killeen, Zeming Lin, Natalia Gimelshein, Luca Antiga, et~al.
\newblock Pytorch: An imperative style, high-performance deep learning library.
\newblock {\em Advances in neural information processing systems},
  32:8026--8037, 2019.

\bibitem{pate1996uncertainties}
M~Elisabeth Pat{\'e}-Cornell.
\newblock Uncertainties in risk analysis: Six levels of treatment.
\newblock {\em Reliability Engineering \& System Safety}, 54(2-3):95--111,
  1996.

\bibitem{pont20172017}
Jordi Pont-Tuset, Federico Perazzi, Sergi Caelles, Pablo Arbel{\'a}ez, Alex
  Sorkine-Hornung, and Luc Van~Gool.
\newblock The 2017 davis challenge on video object segmentation.
\newblock {\em arXiv preprint arXiv:1704.00675}, 2017.

\bibitem{qiao2020deep}
Mu Qiao, Ziyi Meng, Jiawei Ma, and Xin Yuan.
\newblock Deep learning for video compressive sensing.
\newblock {\em APL Photonics}, 5(3):030801, 2020.

\bibitem{Reddy11_CVPR_P2C2}
D. {Reddy}, A. {Veeraraghavan}, and R. {Chellappa}.
\newblock P2c2: Programmable pixel compressive camera for high speed imaging.
\newblock In {\em CVPR 2011}, pages 329--336, June 2011.

\bibitem{simonyan2014very}
Karen Simonyan and Andrew Zisserman.
\newblock Very deep convolutional networks for large-scale image recognition.
\newblock {\em arXiv preprint arXiv:1409.1556}, 2014.

\bibitem{touvron2021training}
Hugo Touvron, Matthieu Cord, Matthijs Douze, Francisco Massa, Alexandre
  Sablayrolles, and Herv{\'e} J{\'e}gou.
\newblock Training data-efficient image transformers \& distillation through
  attention.
\newblock In {\em International Conference on Machine Learning}, pages
  10347--10357. PMLR, 2021.

\bibitem{tsai2015spectral}
Tsung-Han Tsai, Patrick Llull, Xin Yuan, Lawrence Carin, and David~J Brady.
\newblock Spectral-temporal compressive imaging.
\newblock {\em Optics letters}, 40(17):4054--4057, 2015.

\bibitem{tsai2015spatial}
Tsung-Han Tsai, Xin Yuan, and David~J Brady.
\newblock Spatial light modulator based color polarization imaging.
\newblock {\em Optics express}, 23(9):11912--11926, 2015.

\bibitem{tu2022maxvit}
Zhengzhong Tu, Hossein Talebi, Han Zhang, Feng Yang, Peyman Milanfar, Alan
  Bovik, and Yinxiao Li.
\newblock Maxvit: Multi-axis vision transformer.
\newblock {\em arXiv preprint arXiv:2204.01697}, 2022.

\bibitem{wang2021pyramid}
Wenhai Wang, Enze Xie, Xiang Li, Deng-Ping Fan, Kaitao Song, Ding Liang, Tong
  Lu, Ping Luo, and Ling Shao.
\newblock Pyramid vision transformer: A versatile backbone for dense prediction
  without convolutions.
\newblock In {\em Proceedings of the IEEE/CVF International Conference on
  Computer Vision}, pages 568--578, 2021.

\bibitem{wang2018non}
Xiaolong Wang, Ross Girshick, Abhinav Gupta, and Kaiming He.
\newblock Non-local neural networks.
\newblock In {\em Proceedings of the IEEE conference on computer vision and
  pattern recognition}, pages 7794--7803, 2018.

\bibitem{wang2004image}
Zhou Wang, Alan~C Bovik, Hamid~R Sheikh, and Eero~P Simoncelli.
\newblock Image quality assessment: from error visibility to structural
  similarity.
\newblock {\em IEEE transactions on image processing}, 13(4):600--612, 2004.

\bibitem{wang2021metasci}
Zhengjue Wang, Hao Zhang, Ziheng Cheng, Bo Chen, and Xin Yuan.
\newblock Metasci: Scalable and adaptive reconstruction for video compressive
  sensing.
\newblock In {\em Proceedings of the IEEE/CVF Conference on Computer Vision and
  Pattern Recognition}, pages 2083--2092, 2021.

\bibitem{wright1999bayesian}
WA Wright.
\newblock Bayesian approach to neural-network modeling with input uncertainty.
\newblock {\em IEEE Transactions on Neural Networks}, 10(6):1261--1270, 1999.

\bibitem{Wu_2021_ICCV}
Zhuoyuan Wu and etc.
\newblock Dense deep unfolding network with 3d-cnn prior for snapshot
  compressive imaging.
\newblock In {\em ICCV 2021}.

\bibitem{9428320}
Zhuoyuan Wu, Zhenyu Zhang, Jiechong Song, and Man Zhang.
\newblock Spatial-temporal synergic prior driven unfolding network for snapshot
  compressive imaging.
\newblock In {\em 2021 IEEE International Conference on Multimedia and Expo
  (ICME)}, pages 1--6, 2021.

\bibitem{Xu_2020_CVPR}
Ke Xu, Xin Yang, Baocai Yin, and Rynson~W.H. Lau.
\newblock Learning to restore low-light images via
  decomposition-and-enhancement.
\newblock In {\em Proceedings of the IEEE/CVF Conference on Computer Vision and
  Pattern Recognition (CVPR)}, June 2020.

\bibitem{Yang14GMMonline}
J. Yang, X. Liao, X. Yuan, P. Llull, D.~J. Brady, G. Sapiro, and L. Carin.
\newblock Compressive sensing by learning a {G}aussian mixture model from
  measurements.
\newblock {\em IEEE Transaction on Image Processing}, 24(1):106--119, January
  2015.

\bibitem{Yang14GMM}
J. Yang, X. Yuan, X. Liao, P. Llull, G. Sapiro, D.~J. Brady, and L. Carin.
\newblock Video compressive sensing using {G}aussian mixture models.
\newblock {\em IEEE Transaction on Image Processing}, 23(11):4863--4878,
  November 2014.

\bibitem{yin2020disentangled}
Minghao Yin, Zhuliang Yao, Yue Cao, Xiu Li, Zheng Zhang, Stephen Lin, and Han
  Hu.
\newblock Disentangled non-local neural networks.
\newblock In {\em European Conference on Computer Vision}, pages 191--207.
  Springer, 2020.

\bibitem{yoshida2018joint}
Michitaka Yoshida, Akihiko Torii, Masatoshi Okutomi, Kenta Endo, Yukinobu
  Sugiyama, Rin-ichiro Taniguchi, and Hajime Nagahara.
\newblock Joint optimization for compressive video sensing and reconstruction
  under hardware constraints.
\newblock In {\em Proceedings of the European Conference on Computer Vision
  (ECCV)}, pages 634--649, 2018.

\bibitem{yuan2021tokens}
Li Yuan, Yunpeng Chen, Tao Wang, Weihao Yu, Yujun Shi, Zi-Hang Jiang,
  Francis~EH Tay, Jiashi Feng, and Shuicheng Yan.
\newblock Tokens-to-token vit: Training vision transformers from scratch on
  imagenet.
\newblock In {\em Proceedings of the IEEE/CVF International Conference on
  Computer Vision}, pages 558--567, 2021.

\bibitem{yuan2016generalized}
Xin Yuan.
\newblock Generalized alternating projection based total variation minimization
  for compressive sensing.
\newblock In {\em 2016 IEEE International Conference on Image Processing
  (ICIP)}, pages 2539--2543. IEEE, 2016.

\bibitem{Yuan2021_SPM}
X. {Yuan}, D.~J. {Brady}, and A.~K. {Katsaggelos}.
\newblock Snapshot compressive imaging: Theory, algorithms, and applications.
\newblock {\em IEEE Signal Processing Magazine}, 38(2):65--88, 2021.

\bibitem{yuan2020plug}
Xin Yuan and etc.
\newblock Plug-and-play algorithms for large-scale snapshot compressive
  imaging.
\newblock In {\em IEEE CVPR 2020}.

\bibitem{Yuan14CVPR}
Xin Yuan, Patrick Llull, Xuejun Liao, Jianbo Yang, David~J. Brady, Guillermo
  Sapiro, and Lawrence Carin.
\newblock Low-cost compressive sensing for color video and depth.
\newblock In {\em IEEE Conference on Computer Vision and Pattern Recognition
  (CVPR)}, pages 3318--3325, 2014.

\bibitem{zhang2022herosnet}
Xuanyu Zhang, Yongbing Zhang, Ruiqin Xiong, Qilin Sun, and Jian Zhang.
\newblock Herosnet: Hyperspectral explicable reconstruction and optimal
  sampling deep network for snapshot compressive imaging.
\newblock In {\em Proceedings of the IEEE/CVF Conference on Computer Vision and
  Pattern Recognition}, pages 17532--17541, 2022.

\bibitem{zhang2018shufflenet}
Xiangyu Zhang, Xinyu Zhou, Mengxiao Lin, and Jian Sun.
\newblock Shufflenet: An extremely efficient convolutional neural network for
  mobile devices.
\newblock In {\em Proceedings of the IEEE conference on computer vision and
  pattern recognition}, pages 6848--6856, 2018.

\bibitem{zhao2016video}
Chen Zhao, Siwei Ma, Jian Zhang, Ruiqin Xiong, and Wen Gao.
\newblock Video compressive sensing reconstruction via reweighted residual
  sparsity.
\newblock {\em IEEE Transactions on Circuits and Systems for Video Technology},
  27(6):1182--1195, 2016.

\bibitem{zheng2021deep}
Siming Zheng, Yang Liu, Ziyi Meng, Mu Qiao, Zhishen Tong, Xiaoyu Yang,
  Shensheng Han, and Xin Yuan.
\newblock Deep plug-and-play priors for spectral snapshot compressive imaging.
\newblock {\em Photonics Research}, 9(2):B18--B29, 2021.

\bibitem{zheng2021super}
Siming Zheng, Chunyang Wang, Xin Yuan, and Huolin~L Xin.
\newblock Super-compression of large electron microscopy time series by deep
  compressive sensing learning.
\newblock {\em Patterns}, 2(7):100292, 2021.

\end{thebibliography}
}

\end{document}